# New Perspectives in Sinographic Language Processing Through the Use of Character Structure*

Yannis Haralambous

Institut Télécom - Télécom Bretagne
Lab-STICC UMR CNRS 6285
yannis.haralambous@telecom-bretagne.eu

**Abstract.** Chinese characters have a complex and hierarchical graphical structure carrying both semantic and phonetic information. We use this structure to enhance the text model and obtain better results in standard NLP operations. First of all, to tackle the problem of graphical variation we define allographic classes of characters. Next, the relation of inclusion of a subcharacter in a characters, provides us with a directed graph of allographic classes. We provide this graph with two weights: semanticity (semantic relation between subcharacter and character) and phoneticity (phonetic relation) and calculate "most semantic subcharacter paths" for each character. Finally, adding the information contained in these paths to unigrams we claim to increase the efficiency of text mining methods. We evaluate our method on a text classification task on two corpora (Chinese and Japanese) of a total of 18 million characters and get an improvement of 3% on an already high baseline of 89.6% precision, obtained by a linear SVM classifier. Other possible applications and perspectives of the system are discussed.

## 1 Introduction

The Chinese script is used mainly in the Chinese and Japanese languages. Chinese characters (or "sinographs") are notorious for their large number (over 84 thousand have been encoded in Unicode [1]) and their complexity (they can have from 1 stroke, like 一, to as many as 64 strokes, like 𠔻). Despite its complexity, the Chinese script is quite efficient, since semantic and phonetic information is stored in stroke patterns, easily recognizable by native readers. In this paper we will deal with a specific kind of stroke pattern, namely those that exist also as stand-alone characters—we call them *subcharacters*. We will study the phonetic similarity (called *phoneticity*) and the semantic relatedness (called *semanticity*) between a character and its subcharacters. After having built a graph of subcharacter inclusions, and attaching various kinds of information to it, we introduce an enhanced NLP task feature model: together with individual characters we use data contained in their subcharacter graphs. Indeed, in sinographic language processing it is customary to combine character-level and word-level processing. The advantage to our approach is that an additional level is added to these two: the level of subcharacters. By exploring this additional level, we go deeper into the inherent structure of sinographic characters—this inherent structure is completely lost in conventional NLP approaches.

* The final publication is available at `http://link.springer.com`.

We believe that this new feature model will prove useful in various branches of statistical NLP. As a first step in that direction, we evaluate our tools by applying them to a text classification task.

### 1.1 Related Work

There have been various attempts at describing sinographs in a systematic way: [2] and [3] use generative grammars, [4] uses a Prolog-like language, [5] uses the biological metaphor of gene theory, [6] describes sinographs as objects with features (the Chaon model), [7] defines a formal language type called planar regular language, [8] and [9] describe the topology of sinographs in XML, [10] uses projections of stroke bounding boxes, and [11] use combinatorics of IDS operators.[1]

The OCR community has also shown a strong interest in the structure of sinographs [12].

As for considering Chinese script as a network, this has been done in several papers, such as [13, 14] (where edges represent components combined in the same sinograph) or [15], where components and sinographs form a bipartite graph; [16], where edges represent sinographs combined into words; [17], dealing with purely phonemic networks. [14] give an example of a small phono-semantic graph (a bipartite graph where edges connect semantic and phonetic components), but do not enter into the calculations of phoneticity and semanticity. *Hanzi Grid* [18, 19] maps component inclusions to relations in ontologies.

Finally the cognitive psychology community is also heavily interested in the (cognitive) processing of phonetic and semantic components [20, 21], and even in the effects of stroke order and radicals on linguistic knowledge [22].

## 2 Definitions

### 2.1 Strokes

A sinograph consists of a number of *strokes*, arranged inside an (imaginary) square according to specific patterns. Strokes can be classified as belonging to 36 *calligraphic stroke classes*. The latter have been encoded in Unicode (table CJK STROKES). Furthermore, strokes are always drawn in a very specific order.

[1] *Ideographic Description Sequences* do not describe sinographs per se, but provide operators ⿰ ⿱ ⿲ ⿳ ⿴ ⿵ ⿶ ⿷ ⿸ ⿹ ⿺ ⿻ for graphically combining existing sinographs in groups of two or three. IDS operators can be arbitrarily nested and have been encoded in Unicode (table IDEOGRAPHIC DESCRIPTION CHARACTERS).

### 2.2 Components, Subcharacters, Radicals

In a manner similar to etymological roots in Western languages, readers of sinographs recognize patterns of strokes, so that the meaning of an unknown sinograph can be identified, more or less effectively, by the stroke patterns it contains.

Frequently appearing patterns of strokes are called *components*. In this paper we will deal only with components that also exist as isolated sinographs. Using the term "character" in the sense of "Unicode character," we will call such components, *subcharacters*. In other words, subcharacters are *components having a Unicode identity*.

Classification of sinographs in dictionaries traditionally uses a set of several hundred subcharacters called *radicals*. We do not discuss radicals in this paper.

### 2.3 Allographic Classes

An important property of components is that they can change shape when combined with other components (for example, 火 becomes 灬 when combined, like in 点). Some of these variant shapes have been encoded in Unicode. Also, some sinographs can have variant forms. In particular, during the Chinese writing reform [23], about 1,753 sinographs obtained simplified shapes (for example, 蔔 became 卜), which are encoded separately in Unicode.

As these variations in shape do not affect semantic or phonetic properties (at least not at the level of statistical language processing), we merge characters and their variants into sets called *allographic classes*. For example, 糸 belongs to class [⽷,⺯,糸,糸,纟] where the two first characters belong to the table of Unicode radicals and the others are graphical variants.

We obtain 18,686 allographic classes, out of which 87.356% are singletons, the highest number of characters per class is 15 and the average is 1.1382.

In the remainder of this paper we will use italics for characters ($c, s, \ldots$) and bold letters for allographic classes ($\mathbf{c}, \mathbf{s}, \ldots$). The term "subcharacter" will mean a character or an allographic class, depending on the context.

### 2.4 Semanticity and Phoneticity

Subcharacters can play a *semantic role* (when one of their meanings is close to one of the meanings of the sinograph) and/or a *phonetic role* (when one of their readings is identical or close, in a given language, to one of the readings of the sinograph). In this paper we will deal with Mandarin and Japanese.

These two properties of subcharacters in relation to characters can be quantified and are then called *semanticity* and *phoneticity* [24].

## 3 Resources

### 3.1 Frequency Lists

First some notation: let $T$ be a sinographic text (or a corpus considered as a single text). A *frequency list* $A$, generated out of $T$, is an $M$-tuple of pairs $(c_i, f_A(c_i))$, where the frequency $f_A(c_i)$ of character $c_i$ is defined as $\frac{\#c_i}{\#T}$, that is the number of occurrences of $c_i$ in $T$ divided by the length of $T$. The $c_i$ must be pairwise different. $A$ is sorted in decreasing order of frequencies: $f_A(c_i) \geq f_A(c_j)$ when $i < j$. If $N \in \mathbb{N}$, let $A_{1\ldots N}$ be the subtuple of the first $N$ pairs.

Let $\mathrm{char}(A)$ be the underlying set of characters of $A$. Let $A_{1\ldots N}$ be the sublist of the $N$ first characters of $A$. Let

$$\mathrm{comchar}_N(A, A') := (\mathrm{char}(A_{1\ldots N}) \cap \mathrm{char}(A')) \cup (\mathrm{char}(A'_{1\ldots N}) \cap \mathrm{char}(A))$$

be the *set of $N$-common characters* between two lists $A$ and $A'$. In other words, among the first $N$ characters of $A$ we take those that also belong to $A'$ and vice versa. We define the *$N$-common coverage factor* between $A$ and $A'$ as:

$$\mathrm{comcov}_N(A, A') := \frac{\#\mathrm{comchar}(A, A')}{N}.$$

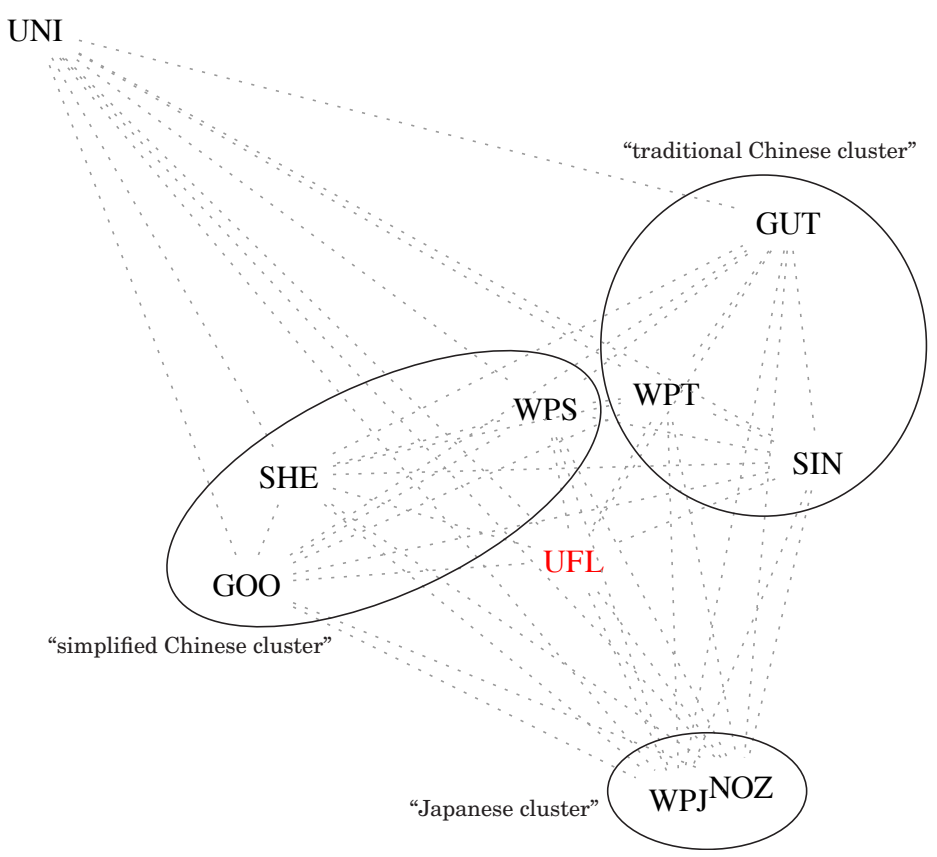

**Fig. 1.** Sinograph frequency lists.

Using $\mathrm{comchar}_N(A, A')$ as the underlying character set, we obtain sub-lists $A_c$ of $A$ and $A'_c$ of $A'$ (although not noted, $A_c$ depends not only on $A$ but also on $A'$ and on $N$, and $A'_c$ also on $A$ and $N$).

Finally, we define a *distance of character frequency lists* $d_N$ as follows:

$$d_N(A, A') := 1 - \mathrm{comcov}_N(A, A') \cdot \frac{\rho(A_c, A'_c) + 1}{2}$$

where $\rho$ is the Spearman ranking correlation coefficient ($\rho$ takes values in $[-1, 1]$).

We have used three publicly available frequency lists: UNI, the language-independent *Unihan* [25]; SIN, the traditional Chinese *Sinica* compiled by Academia Sinica Taipei; and NOZ, a Japanese list, taken from [26].

We have also compiled our own frequency lists out of five corpora: Chinese Wikipedia (WPS for simplified Chinese and WPT for traditional Chinese), Japanese Wikipedia WPJ, Chinese Project Gutenberg GUT, Chinese GoogleBooks GOO, Leeds Chinese Internet Corpus CIC [27].

As we needed a frequency list suitable for all sinographic languages, we calculated distance $d_N$ between them (for $N = 3,000$). In Fig. 1 the reader can see a graph of frequency lists with edges proportional to values of $d_N$. One can identify three linguistic clusters, while the Unihan list can be considered as an outlier. After removing the Unihan list we aggregated the remaining lists to form a "Universal Frequency List," UFL, as the normalized average of NOZ, WPJ, WPS,

WPT, SHE, GOO, SIN and GUT, defined as follows: if $f_X(c)$ is the frequency of character $c$ in corpus $X$, and $\#X$ is the size of the corpus in characters, then:

$$\mathrm{char}(\textsc{ufl}) = \bigcup_{X\in\mathcal{X}} \mathrm{char}(X), \qquad f_{\textsc{ufl}}(c) = \sum_{X\in\mathcal{X}} \frac{f_X(c)\cdot \#\mathrm{char}(X)}{\#\mathrm{char}(\textsc{ufl})},$$

where $\mathcal{X} = \{$NOZ, WPJ, WPS, WPT, SHE, GOO, SIN, GUT$\}$.

### 3.2 Character Descriptions and Subcharacter Inclusions

Wenlin Institute kindly provided us with the CDL database of sinographs. This XML file provides an ordered stroke list for each sinograph, and for each stroke, its calligraphic type and coordinates of endpoints. We modified stroke order for the few exceptional cases where components are overlapping.[2]

Because of the many affine transformations a component is subject to, this topological sinograph description is not suitable for effectively detecting subcharacters. On the other hand, use of topological properties is unavoidable, since some sinographs have the same strokes in the same order and combinatorial arrangement but differ by the (relative) size of strokes, the typical example being 士 (scholar) and 土 (earth), where the bottom stroke of the latter is longer than that of the former. For this reason we used a different representation of sinographs, based on relative size of strokes and extrapolated intersection locations. For example, here is the relation between the two first strokes of the sinograph 言, which are of type d (= "dot") and h (= "horizontal"):

```
d
(1.4,9.1,0.0,0.6)
h
```

(see Fig. 2 for the description of the numeric values).

Using this affine transformation invariant representation of sinographs, we extracted 868,856 subcharacter inclusions from the Wenlin CDL database, where inclusion $s \to c$ of subcharacter $s$ into character $c$ means that the code lines of our representation of $s$ are contained, in their identical form, in the representation of $c$. After merging with data from the CHISE project [28] and with data kindly provided by Cornelia Schindelin [29], and after having removed identities, we got a list of 824,120 strict inclusions. The number is quite high because inclusion chains like: 丿 $\to$ 勹 $\to$ 匐 $\to$ 葍 provide automatically all triangulations 丿 $\to$ 匐, 勹 $\to$ 葍, etc., and there are exponentially many of them. We have detriangulated by taking systematically the longest path, and hence reduced the number of inclusions to 185,801.

---

[2] An important fact about components is that in the sequence of strokes drawn in traditional stroke order, *components form subsequences without overlapping*: the first stroke of a component is drawn after the last stroke of the preceding component. There are a few exceptions to this rule: sinographs like 团 (subcharacter 囗 containing 才) where the drawing of the lower horizontal stroke of 囗 is postponed until the drawing of the internal subcharacter 才 has been completed.

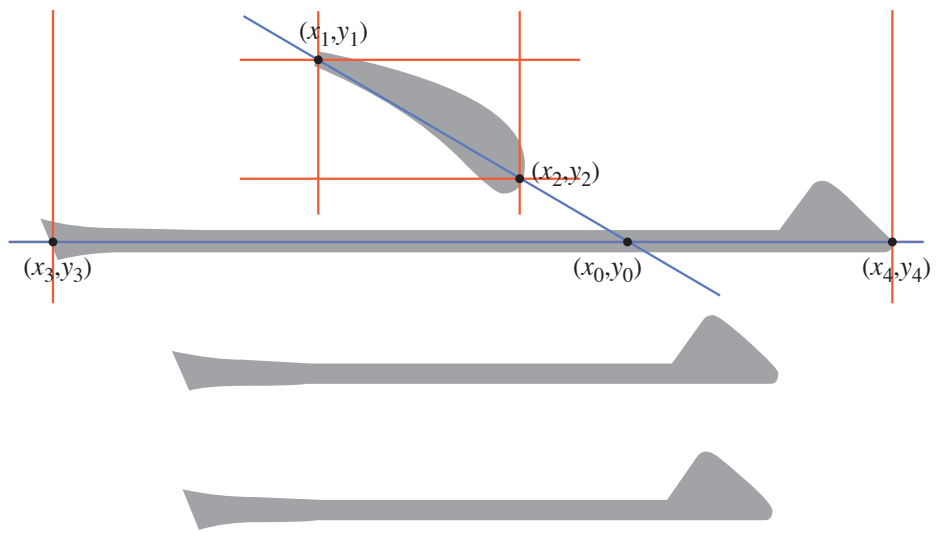

**Fig. 2.** Representation of sinographs based on relative sizes of strokes and extrapolated intersection locations. The blue lines are skeletons of strokes given by CDL. The values `(1.4,9.1,0.0,0.6)` correspond to: $\frac{d((x_0,y_0)-(x_1,y_1))}{d((x_2,y_2)-(x_1,y_1))}$ (that is: distance of the intersection point from $(x_1, y_1)$ with stroke length as unit), $\frac{|x_4-x_3|}{|x_2-x_1|}$ (stroke box width ratio), $\frac{|y_4-y_3|}{|y_2-y_1|}$ (stroke box height ratio) and $\frac{d((x_0,y_0)-(x_3,y_3))}{d((x_4,y_4)-(x_3,y_3))}$ (again, distance of the intersection point from $(x_3, y_3)$ with second stroke length as unit), where $d$ is Euclidean distance. When strokes are parallel or orthogonal, some of the parameters are infinite and we write `E` instead.

### 3.3 The Inclusion Graph

We construct a graph, using all sinographs as vertices and representing the inclusions as edges, giving us 74,601 vertices and 185,801 edges. Both the in-degree and out-degree properties of this graph (see Fig. 3) follow a power law distribution of parameters $\alpha^- = 1.138$ (in-degree) and $\alpha^+ = 1.166$ (out-degree). Remarkably, these are a bit low compared to typical scale-free networks like the Web or proteins, which are in the 2–3 range [30].

As higher Unicode planes contain rare sinographs which are of little use to common NLP tasks, and for reasons of computational efficiency, we restricted ourselves to the subgraph of sinographs contained in Unicode's BMP (Basic Multilingual Plane). This subset covers only 91.65% of the sinographs contained in the frequency list, but its frequency-weighted coverage[3] is as high as 99.9995%, showing that our choice of BMP is justified.

By lifting[4] sinograph inclusions to allographic classes we obtain a graph $\mathcal{C}$ of 18,686 allographic classes and 39,719 class inclusions.

In all, 99.8% of the allographic classes have incoming inclusions (the highest in-degree is 12 for class [華]), 14 classes are "sources" (zero in-degree nodes): [一], [丨], [丶], [丿], [亅], [力], [巛], [乀, ⺄, 乛, 乚, 乙], [乃], [刁], [又], [及], [巜] and [廴]. 87.05% of the classes are leaves (zero out-degree nodes) and the highest out-degree is 996 for class [氵, 水]).

[3] If standard coverage is $\frac{\#\{c_i \in \mathrm{char}(\textsc{ufl})\}}{\#\{c_i \in \mathrm{BMP}\}}$ where UFL is our Universal Frequency List, then *frequency-weighted coverage* is defined as $\frac{\sum_{c_i \in \mathrm{char}(\textsc{ufl})} f_{\textsc{ufl}}(c_i)}{\sum_{c_i \in \mathrm{BMP}} f_{\textsc{ufl}}(c_i)}$.

[4] If $\mathbf{c}_1, \mathbf{c}_2 \in \mathcal{C}$ we have $\mathbf{c}_1 \to \mathbf{c}_2$ iff there is at least one character pair $c_1 \in \mathbf{c}_1, c_2 \in \mathbf{c}_2$ such that $c_1 \to c_2$.

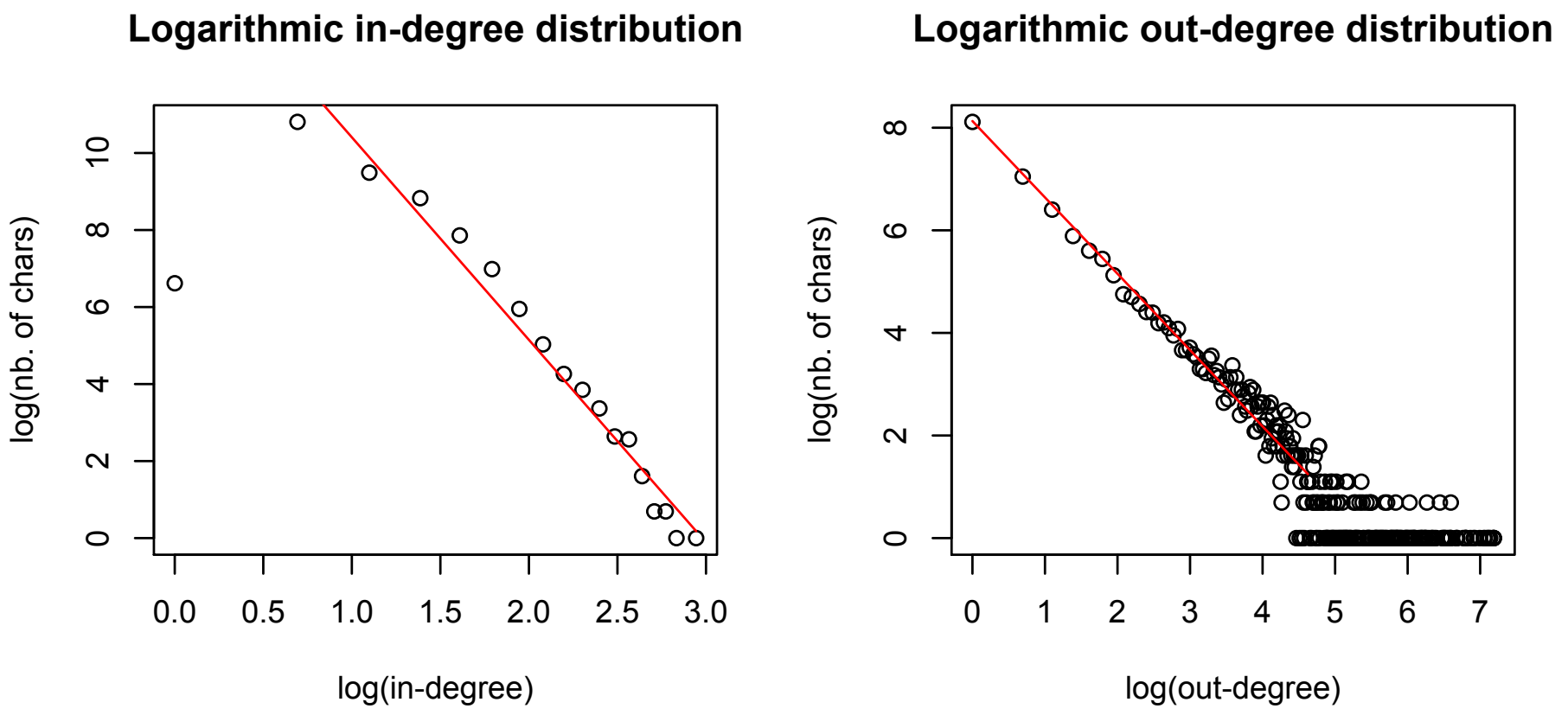

**Fig. 3.** Power-law fitting of the in- and out-degree distribution of the graph of allographic classes.

## 4 Phoneticity

Unihan provides phonetic data for sinographs in several languages. For this study, we used data for Mandarin Chinese, Japanese On (readings originating from China, and imported to Japan together with the writing system) and Japanese Kun (native Japanese readings). We define *phoneticity* as the degree of phonetic similarity between subcharacter and character. To calculate it we need a phonetic distance.

For Mandarin Chinese we implemented the phonetic distance described in [31].

For Japanese we have defined our own phonetic distance.We obtained this distance by applying a methodology given by [32] and [33]: indeed, we defined a distance for syllables, using seven features: consonant place of articulation (weight 4), consonant voicing (1), consonant manner of articulation (4), consonant palatalization (1), vowel frontness (5), vowel height (1), and vowel rounding (1). Distance between syllables is Euclidean in feature space. When sinograph readings have an unequal number of syllables[5] we use a sliding window approach to find the shortest phonetic distance between the shorter word and a subword of equal length to the longer one.

If $d$ is a phonetic distance between sinographs, let $d_{\min}$ be the distance between allographic classes defined as the minimum distance between class members. We define the *phoneticity coefficient* $\varphi(\mathbf{c}_1, \mathbf{c}_2) := \frac{1-d_{\min}(\mathbf{c}_1,\mathbf{c}_2)}{N}$ where $N$ is a normalization constant such that $\mathrm{Im}(\varphi) = [0, 1]$. We will write $\varphi(\mathbf{s} \to \mathbf{c})$ for $\varphi(\mathbf{s}, \mathbf{c})$ when $\mathbf{s} \to \mathbf{c}$ is a class inclusion. In Fig. 4, we show the distribution density of $\varphi$ for Mandarin, Japanese On and Kun. One can see that On mimics the distribution of Mandarin (with a lesser $\varphi$ value for the right peak) while Kun has a completely different distribution with a single peak around $\varphi = 0.4$. Indeed, the historical relation of On and Mandarin is reflected in the similarity of distributions, while in Kun phonetic distance between subcharacter and character is random.

[5] Contrary to Chinese language where sinograph readings are monosyllabic, in Japanese they can have up to 12 syllables (the longest readings are *nuhitorioshitanameshigaha* for 韇 and *hitohenotsutsusodeudenuki* for 褠, both being Kun readings).

Here is an example:

| | Mandarin | Jap. On | Jap. Kun |
|---|---|---|---|
| 任 | rèn | nin | makaseru, ninau, taeru |
| $\uparrow$ | $\uparrow\approx$ | $\uparrow=$ | $\uparrow\neq$ |
| 人 | rén | nin | hito |

where the difference between the phonetics of the two characters in Mandarin is at the tone level only, and since tones are not phonemic in Japanese, the sinographs are homophones in On. Incidentally, this inclusion has very low semanticity: 人 means "man" and 任, "to trust."

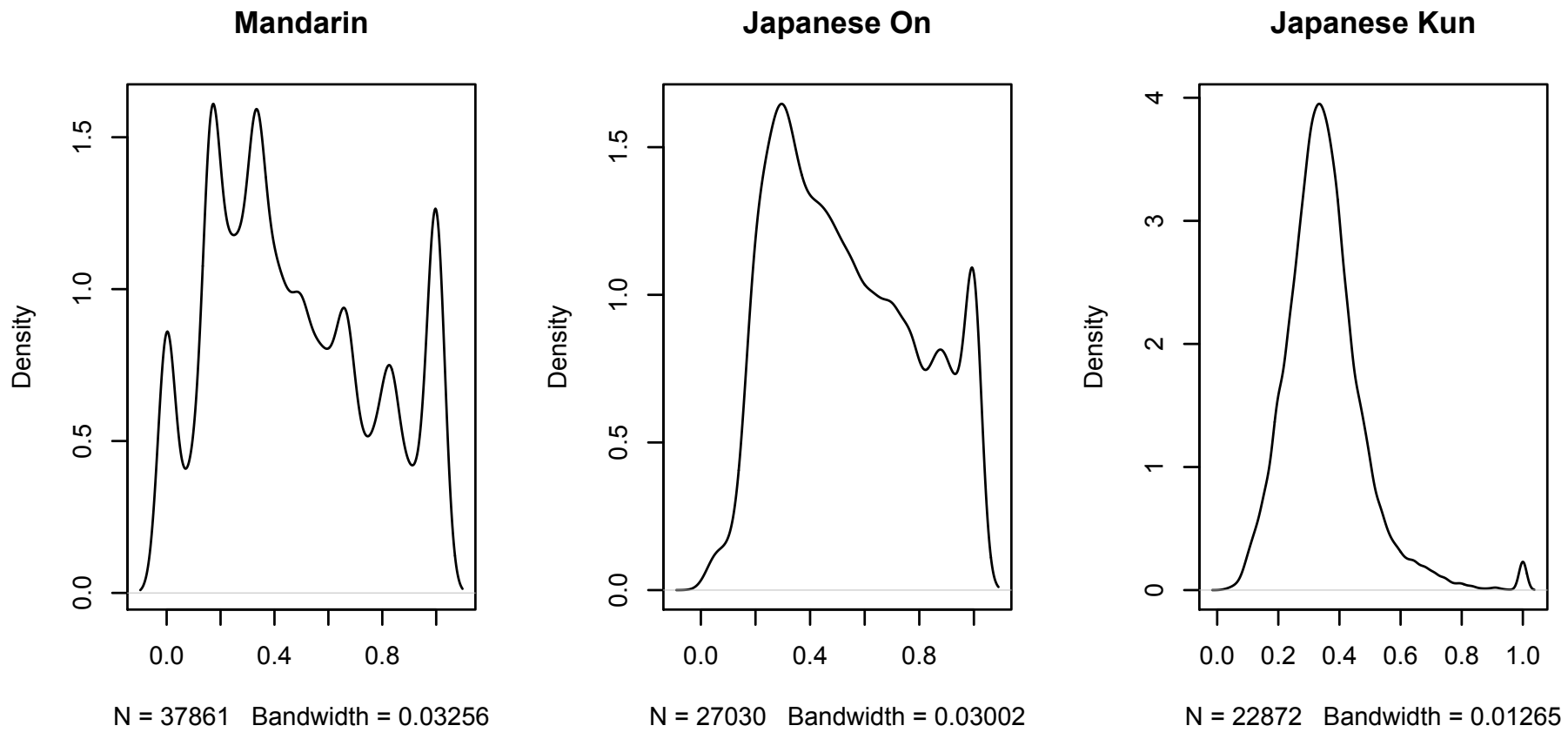

**Fig. 4.** Phoneticity coefficient distribution for Mandarin, Japanese On and Kun.

### 4.1 The Least Phonetic Chain

Under the hypothesis that *subcharacters with higher phoneticity have statistically lower semanticity and vice versa*, we consider the subcharacters with the lowest phoneticity, as these have an increased potential of having higher semanticity. We define the *least phonetic chain* $(\mathbf{p}_i)_{i\geq 0}$ of a class $\mathbf{c}$ as follows: let $\mathbf{p}_0 = \mathbf{c}$ and given $\mathbf{p}_i$ let $\mathbf{p}_{i+1} := \mathrm{argmin}_{\mathbf{z}}\varphi(\mathbf{z} \to \mathbf{p}_i)$, that is, the subcharacter of $\mathbf{p}_i$ with least phoneticity (see Fig. 5 for an example). We will use this construct in our text classification strategies.

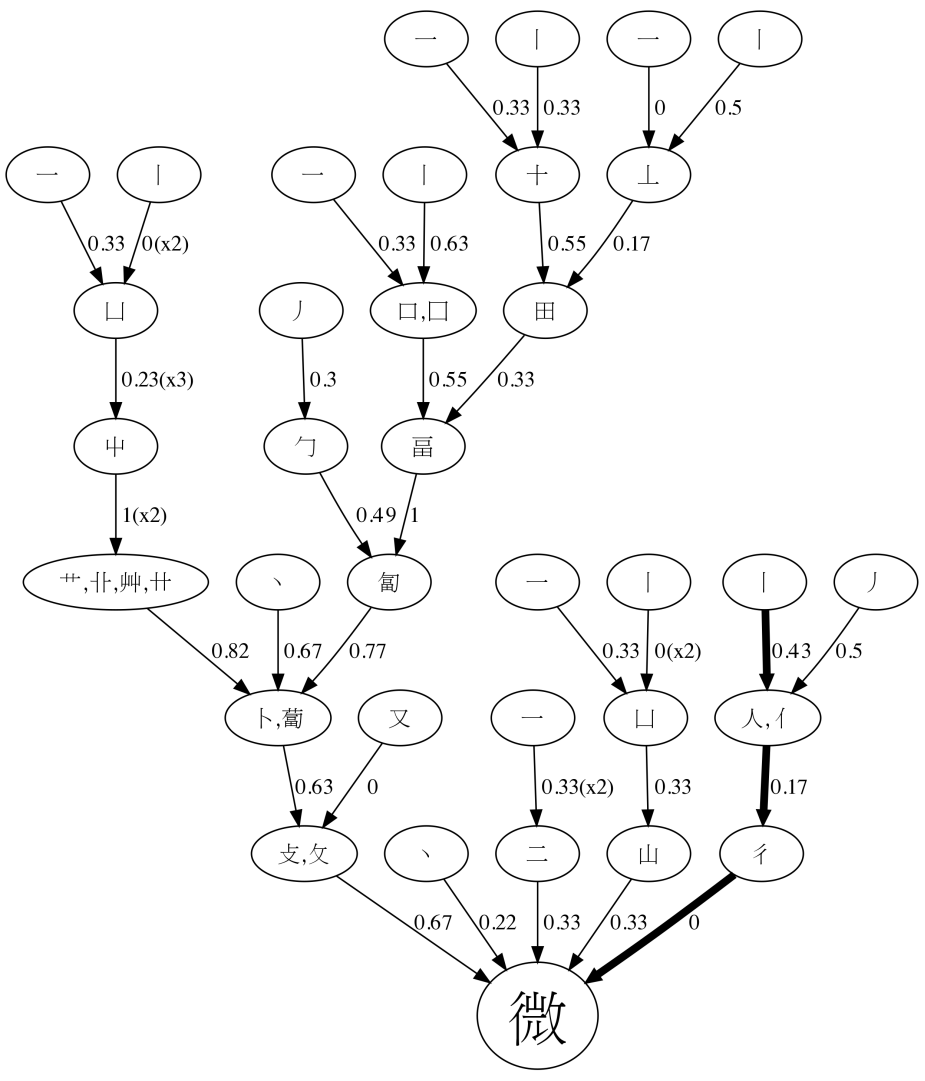

**Fig. 5.** The inclusion tree of allographic class [微], in bold: the least phonetic chain. Edge labels are $\varphi$ values; between parentheses, the subcharacter's multiplicity.

## 5 Semanticity

As semantic resources for sinographic languages we used three WordNets: the Academia Sinica BOW [34] for traditional Chinese, the Chinese WordNet [35] for simplified Chinese and the Japanese WordNet [36]. All three provide English WordNet synset IDs. The single-sinograph entries they contain are rather limited: 3,075 for traditional Chinese, 2,440 for simplified Chinese and 4,941 for Japanese. From these we obtain a first mapping of allographic classes into synset IDs. In all, 2,852 allographic classes are covered, out of which 1,063 have a single synset ID, 581 have two IDs (IDs from different WordNets are not merged), etc. The highest number of IDs is 45, for [刺] (= "thorn"). Only 154 classes share the same synset ID in all three WordNets.

The Unihan database also provides meanings for a large number of sinographs, but these meanings are not mapped to WordNet. We used the following method to attach synset IDs to them: let $w_i$ be a Chinese or Japanese word attached to synset ID $\sigma_i$ in one of the three WordNets; let $e_{i,k}$ be one of the terms of the English WordNet synset with ID $\sigma_i$; if $e_{i,k}$ can be found in Unihan as meaning of a character $c_j$ *and* if $c_j \in w_i$ then we attach synset ID $\sigma_i$ to $c_j$. We lift this information to allographic classes.

Here is an example:

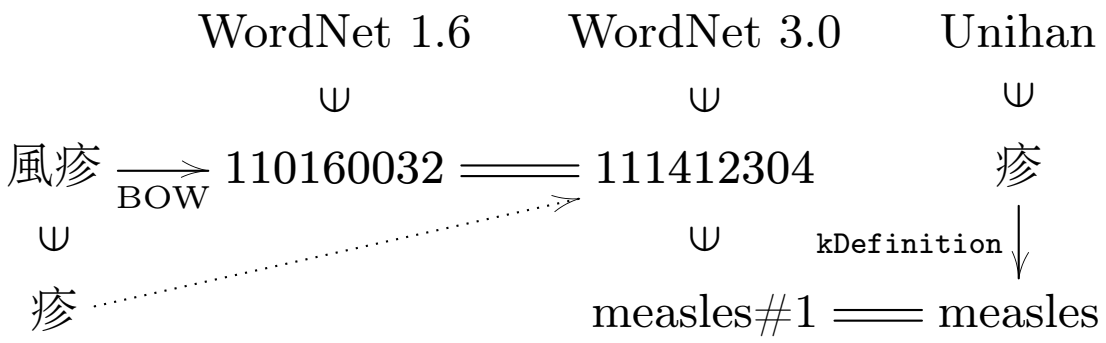

Since 疹 has the meaning "measles" in Unihan and is contained in the word 風疹 which belongs to the measles synset, we attach the measles synset ID to 疹.

Thanks to this method, 1,392 additional allographic classes were mapped to at least one WordNet synset ID, raising the number of semantically annotated classes to 4,244.

### 5.1 Extracting Semantic Relations

We attempted three methods of estimating semanticity of subcharacters:

1. By measuring distance between WordNet nodes using semantic similarity measures, such as those of [37–39] or simply the inverse of the shortest path length in the WordNet graph. This method was unfruitful.

2. By using the following algorithm: whenever there was a semantic relation (hyponymy, meronymy, antonymy, etc.) between two WordNet synsets $\sigma_1$ and $\sigma_2$, and we have a sinograph belonging to a word from $\sigma_2$ and one of its subcharacters in a word from $\sigma_1$, we added a unit of "semantic weight" to the inclusion. We counted these over the three WordNets and obtained 6,816 allographic class inclusions of various weights.

Here is an example:

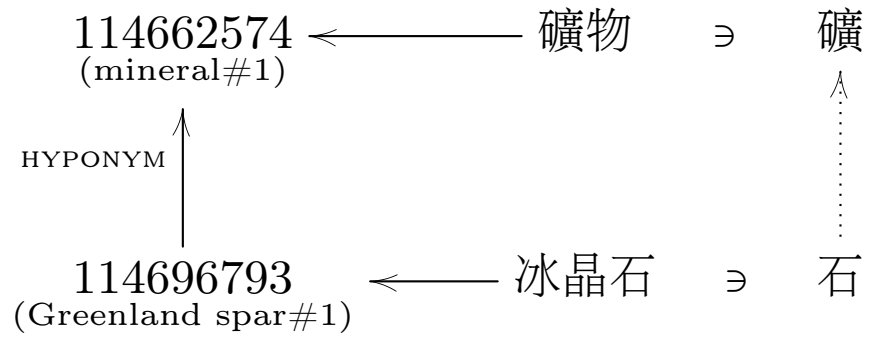

"Greenland spar" is an hyponym of "mineral," 冰晶石 belongs to the synset of the former and 礦物 to that of the latter. Characters 石 and 礦 appear in these two words, and the former being a subcharacter of the latter, we attach a unit of "semantic weight" to it. The (log of the) total amount of units is our tentative semanticity measure.

3. To a lesser degree, we used the Kāng Xī radical (see next section) as an additional semanticity indicator. We did the following: for every inclusion, we compared the Kāng Xī radicals on both sides: when equal, then this may be a hint that the given subcharacter has higher semanticity than others.

Example: in the inclusion 每 → 毓, both sinographs have the same Kāng Xī radical 毋, so we assume that 每 → 毓 has higher semanticity than 㐬 → 毓 (which, incidentally, also has higher phoneticity since it is pronounced *liú* which is closer to *yù* (毓) than *měi* (每)).

# 6 Evaluation

To evaluate an application of our graph to a common NLP task we attempted text classification on two corpora:

1. The Sogou Corpus of Chinese news [43], a collection of 2.36 million online articles (611 million sinographs) taken from various sources. We removed all sinographs not in the CJK UNIFIED IDEOGRAPHS Unicode table and extracted 5,000 texts per class:

| Category | Avg length | Min length | Max length |
|---|---|---|---|
| Sports | 728.23 | 500 | 1,000 |
| Finance | 714.51 | 500 | 1,000 |
| News | 699.63 | 500 | 1,000 |
| Entertainment | 696.36 | 500 | 1,000 |
| Global | 709.68 | 500 | 1,000 |

2. A corpus we built from the online archives of Japanese Reuters [44], covering the period from 2007 to today. We removed all sinographs not in the CJK CHARACTERS Unicode table (including the kana) and extracted 5,000 texts per class. Because of the nature of Reuters news, these texts are shorter than the Chinese ones:

| Category | Avg length | Min length | Max length |
|---|---|---|---|
| Sports | 105.54 | 74 | 521 |
| Finance | 293.79 | 188 | 1,262 |
| News | 315.69 | 166 | 876 |
| Entertainment | 94.73 | 44 | 588 |
| Global | 202.44 | 44 | 1,262 |

Our baseline is obtained as follows: we take unigrams for all sinographs appearing at least 10 times in the corpus. This results in 4,275 unigrams for the Chinese and 2,010 for the Japanese corpus. Then we apply a linear SVM with 10-fold cross validation. Here are the results obtained:

| | Accuracy | # of SVs |
|---|---|---|
| Baseline Chinese | 89.605% | 4,933 |
| Baseline Japanese | 86.925% | 4,237 |

## 6.1 Strategy 1: The Most Semantic Chain

Let $\iota : \mathbf{s} \to \mathbf{c}$ be a class inclusion. We calculate three quantities:

1. $f_1(\iota)$: the frequency of character pairs $(s \in \mathbf{s}, c \in \mathbf{c})$ contained in words $s \in w_1, c \in w_2$ belonging to synsets $w_1 \in \Sigma_1, w_2 \in \Sigma_2$ such that there is a semantic relation $\Sigma_1 \to \Sigma_2$ in WordNet (see Section 5);
2. $f_2(\iota)$: the frequency of character pairs $(s \in \mathbf{s}, c \in \mathbf{c})$ contained in words $s \in w_1, c \in w_2$ belonging to synsets $w_1 \in \Sigma_1, w_2 \in \Sigma_2$ such that there is a two-step semantic relation $\Sigma_1 \to \Sigma' \to \Sigma_2$ in WordNet;
3. $r(\iota)$: let $r(s, c)$ be 1 when $s$ and $c$ share the same Kāng Xī radical, and 0 otherwise. $r(\iota)$ will be $\frac{1}{nm} \sum_{s\in\mathbf{s}} \sum_{c\in\mathbf{c}} r(s, c)$.

The semanticity $S(\iota)$ of inclusion $\iota : \mathbf{s} \to \mathbf{c}$ will be:

$$S(\iota) = \frac{1}{2}\log(1 + f_1(\iota)) + \frac{1}{4}\log(1 + f_2(\iota)) + \frac{1}{4}r(\iota),$$

normalized so that its values stay in the interval $[0, 1]$. Here, coefficients $\frac{1}{2}$, $\frac{1}{4}$ and $\frac{1}{4}$ have been obtained heuristically by a grid method applied to a series of tests.

The *most semantic chain* $(\mathbf{s}_i)_{i\geq 0}$ of class $\mathbf{c}$ is calculated as follows: $\mathbf{s}_0 = \mathbf{c}$, and, given $\mathbf{s}_i$, $\mathbf{s}_{i+1} := \operatorname{argmax}_{\mathbf{z}} S(\mathbf{z} \to \mathbf{s}_i)$, that is, the subcharacter of $\mathbf{s}_i$ with maximal semanticity.

Let $w(c)$ be the weight of unigram $c$, obtained by frequency in the baseline case. Let $w(\mathbf{c}) := \max_{c\in\mathbf{c}} w(c)$ be the weight of class $\mathbf{c}$. For every member $\mathbf{s}_i$ of the most semantic chain of $\mathbf{c}$, we added a weight $w(\mathbf{s}_i) = \frac{1}{i}S(\mathbf{s}_{i-1} \to \mathbf{s}_i)$ (with $\mathbf{s}_0 = \mathbf{c}$) to unigram $c$.

We obtained the following results:

| | Accuracy | # of SVs | Un.w/ mod. | Un.added |
|---|---|---|---|---|
| Chinese | **92.62%** | 3,287 | 594 | 20 |
| Japanese | 89.99% | 3,728 | 524 | 152 |

where the two last columns contain the number of updated unigrams and the number of new unigrams. Notice that the increase in performance is similar for Chinese and Japanese, while the number of support vectors needed is higher in Japanese, probably due to the shorter length of texts making the classification task harder.

### 6.2 Strategy 2: Combining Most Semantic Chain and Least Phonetic Chain

With the notation of previous sections, let $\varphi(\mathbf{s} \to \mathbf{c})$ be the phoneticity of inclusion $\mathbf{s} \to \mathbf{c}$ (calculated as explained in Section 4). Recall that the least phonetic chain $(\mathbf{p}_i)_{i\geq 0}$ is obtained by taking $\mathbf{p}_0 = \mathbf{c}$, and, given $\mathbf{p}_i$, $\mathbf{p}_{i+1} := \operatorname{argmin}_{\mathbf{z}} \varphi(\mathbf{z}, \mathbf{p}_i)$. For each $\mathbf{p}_i$ we define a new class weight $w'(\mathbf{p}_i)$ as follows:

$$w'(\mathbf{p}_i) = w(\mathbf{p}_i) + \frac{1}{i}\varphi(\mathbf{p}_{i-1} \to \mathbf{p}_i).$$

We obtained the following results:

| | Accuracy | # of SVs | Un.w/ mod. | Un.added |
|---|---|---|---|---|
| Chinese | 92.435% | 3,299 | 851 | 245 |
| Japanese | **90.125%** | 3,737 | 745 | 453 |

The value for Chinese is a bit lower than in Strategy 1, but we get a better result for Japanese. We have an increase in the number of new unigrams, due to the fact that we have significantly more phonetically annotated inclusions than semantically annotated ones.

## 7 Unknown Character Semantic Approximation

Besides the usual NLP tasks where a semantic relatedness distance is needed, our system can also be applied to the processing of unknown sinographs. Let $u$ be an unknown sinograph (*i.e.*, we know neither reading nor meaning), and $\mathbf{u}$ its class. In our graph, there is subgraph $G_{\mathbf{u}}$ generated by paths leading to $\mathbf{u}$. The way the graph is built, our subgraph has no cycles. When we leave $\mathbf{u}$ and head towards the leaves, for every path $\mathbf{s}_i$ there will be a first node $\mathbf{n}_i$ for which we have semantic annotation (for example, a WordNet synset ID). Furthermore, we can attach weights to the $\mathbf{n}_i$ by using the product of semanticity expectations (calculated in a way similar to phoneticity expectations) of the edges of path $\mathbf{un}_i$ and the distance from $\mathbf{u}$ on the path $\mathbf{un}_i$.

We obtain a vector in the space of WordNet synset IDs. This is not a precise semantic, but it can be used in various statistical NLP tasks.[6]

## 8 Perspectives

Further work will involve three main areas of focus:

1. further analyzing the graph and extracting knowledge about sinographic languages;
2. similarly processing various higher-order graphs ($n$-grams, words, concepts, ...) and studying their interactions;
3. last, but not least, in the frame of the HanziGraph Project, manually validating the semanticity of the 39,719 class inclusions of our graph by a team of native Chinese/Japanese speakers, using the dedicated Web site `http://www.hanzigraph.net`.

[6] See also [45] where radicals are used to guess not semantics but rather a sinograph's POS tag.

||あ|い|う|え|お|や|ゆ|よ|か|き|く|け|こ|きゃ|きゅ|きょ|さ|し|す|せ|そ|しゃ|しゅ|しょ|た|ち|つ|て|と|ちゃ|ちゅ|ちょ|な|に|ぬ|ね|の|にゃ|にゅ|にょ|は|ひ|ふ|へ|ほ|ひゃ|ひゅ|ひょ|ま|み|む|め|も|みゃ|
|---|---|---|---|---|---|---|---|---|---|---|---|---|---|---|---|---|---|---|---|---|---|---|---|---|---|---|---|---|---|---|---|---|---|---|---|---|---|---|---|---|---|---|---|---|---|---|---|---|---|---|---|---|---|---|
|あ|0|.55|.59|.35|.46|.58|.82|.74|.43|.70|.73|.56|.63|.48|.76|.67|.35|.65|.68|.49|.58|.94|.94|.94|.25|.60|.64|.43|.53|.93|.93|.93|.50|.74|.77|.61|.68|.55|.80|.71|.63|.83|.86|.72|.78|.99|.99|.99|.25|.60|.64|.43|.53|.33|
|い||0|.53|.20|.56|.80|.79|.80|.70|.43|.69|.48|.71|.73|.72|.74|.65|.35|.64|.41|.66|.73|.73|.73|.60|.25|.59|.32|.61|.69|.69|.69|.74|.50|.73|.54|.75|.77|.76|.78|.83|.63|.82|.66|.84|.85|.85|.85|.60|.25|.59|.32|.61|.64|
|う|||0|.46|.16|.82|.58|.60|.73|.69|.43|.63|.46|.76|.48|.51|.68|.64|.35|.58|.39|.93|.93|.93|.64|.59|.25|.52|.30|.91|.91|.91|.77|.73|.50|.68|.53|.80|.55|.57|.86|.82|.63|.78|.65|.99|.99|.99|.64|.59|.25|.52|.30|.68|
|え||||0|.43|.68|.74|.72|.56|.48|.63|.43|.61|.60|.67|.65|.49|.41|.58|.35|.55|.80|.80|.80|.43|.32|.52|.25|.49|.78|.78|.78|.61|.54|.68|.50|.66|.65|.71|.69|.72|.66|.78|.63|.76|.90|.90|.90|.43|.32|.52|.25|.49|.48|
|お|||||0|.74|.60|.58|.63|.71|.46|.61|.43|.67|.51|.48|.58|.66|.39|.55|.35|.95|.95|.95|.53|.61|.30|.49|.25|.93|.93|.93|.68|.75|.53|.66|.50|.71|.57|.55|.78|.84|.65|.76|.63|.99|.99|.99|.53|.61|.30|.49|.25|.57|
|や||||||0|.59|.46|.31|.63|.67|.47|.56|.38|.70|.60|.31|.63|.66|.46|.55|.93|.93|.93|.36|.65|.69|.50|.59|.94|.94|.94|.35|.65|.68|.49|.58|.41|.72|.62|.29|.62|.66|.45|.55|.93|.93|.93|.41|.68|.72|.54|.62|.47|
|ゆ|||||||0|.16|.67|.62|.31|.55|.35|.70|.38|.42|.66|.62|.31|.55|.35|.92|.92|.92|.69|.64|.36|.58|.40|.93|.93|.93|.68|.64|.35|.58|.39|.72|.41|.44|.66|.61|.29|.54|.33|.92|.92|.92|.72|.67|.41|.61|.44|.75|
|よ||||||||0|.56|.64|.35|.53|.31|.60|.42|.38|.55|.64|.35|.53|.31|.94|.94|.94|.59|.67|.40|.56|.36|.95|.95|.95|.58|.66|.39|.55|.35|.62|.44|.41|.55|.63|.33|.52|.29|.94|.94|.94|.62|.69|.44|.59|.41|.66|
|か|||||||||0|.55|.59|.35|.46|.22|.63|.51|.27|.61|.65|.44|.54|.93|.93|.93|.25|.60|.64|.43|.52|.93|.93|.93|.25|.60|.64|.43|.53|.33|.68|.57|.21|.58|.62|.40|.51|.92|.92|.92|.40|.67|.71|.53|.61|.45|
|き||||||||||0|.53|.20|.56|.59|.58|.60|.61|.27|.60|.34|.62|.70|.70|.70|.60|.25|.59|.32|.61|.69|.69|.69|.60|.25|.59|.32|.61|.64|.63|.65|.58|.21|.57|.29|.60|.68|.68|.68|.67|.40|.67|.45|.68|.71|
|く|||||||||||0|.46|.16|.63|.22|.27|.65|.60|.27|.53|.32|.91|.91|.91|.64|.59|.25|.52|.30|.91|.91|.91|.64|.59|.25|.52|.30|.68|.33|.37|.62|.57|.21|.50|.26|.90|.90|.90|.71|.67|.40|.60|.43|.74|
|け||||||||||||0|.43|.41|.51|.48|.44|.34|.53|.27|.51|.78|.78|.78|.43|.32|.52|.25|.49|.77|.77|.77|.43|.32|.52|.25|.49|.48|.56|.54|.40|.29|.50|.21|.47|.77|.77|.77|.53|.45|.60|.40|.58|.57|
|こ|||||||||||||0|.51|.27|.22|.54|.62|.32|.51|.27|.94|.94|.94|.52|.61|.30|.49|.25|.93|.93|.93|.53|.61|.30|.49|.25|.57|.37|.33|.51|.60|.26|.47|.21|.93|.93|.93|.61|.68|.43|.58|.40|.65|
|きゃ||||||||||||||0|.59|.46|.35|.65|.68|.49|.58|.94|.94|.94|.33|.64|.67|.48|.57|.94|.94|.94|.33|.64|.68|.48|.57|.25|.64|.53|.30|.62|.66|.46|.55|.93|.93|.93|.45|.71|.74|.57|.65|.40|
|きゅ|||||||||||||||0|.16|.68|.64|.35|.57|.38|.93|.93|.93|.67|.63|.33|.56|.37|.92|.92|.92|.68|.63|.33|.56|.37|.64|.25|.30|.66|.61|.30|.55|.34|.92|.92|.92|.74|.70|.45|.64|.48|.71|
|きょ||||||||||||||||0|.58|.66|.38|.55|.35|.95|.95|.95|.57|.65|.37|.54|.33|.95|.95|.95|.57|.65|.37|.54|.33|.53|.30|.25|.55|.63|.34|.52|.30|.94|.94|.94|.65|.72|.48|.62|.45|.61|
|さ|||||||||||||||||0|.55|.59|.35|.46|.91|.91|.91|.11|.56|.60|.36|.47|.91|.91|.91|.44|.70|.73|.56|.64|.49|.77|.67|.38|.66|.70|.51|.60|.95|.95|.95|.31|.63|.67|.47|.56|.38|
|し||||||||||||||||||0|.53|.20|.56|.65|.65|.65|.56|.11|.55|.23|.57|.66|.66|.66|.70|.44|.69|.48|.71|.73|.73|.74|.66|.38|.65|.43|.67|.74|.74|.74|.63|.31|.62|.37|.64|.67|
|す|||||||||||||||||||0|.46|.16|.89|.89|.89|.60|.55|.11|.47|.20|.89|.89|.89|.73|.69|.44|.63|.47|.77|.49|.52|.70|.65|.38|.59|.41|.93|.93|.93|.67|.62|.31|.55|.35|.70|
|せ||||||||||||||||||||0|.43|.74|.74|.74|.36|.23|.47|.11|.44|.75|.75|.75|.56|.48|.63|.44|.61|.60|.67|.65|.51|.43|.59|.38|.57|.81|.81|.81|.47|.37|.55|.31|.53|.52|
|そ|||||||||||||||||||||0|.92|.92|.92|.47|.57|.20|.44|.11|.92|.92|.92|.64|.71|.47|.61|.44|.67|.52|.49|.60|.67|.41|.57|.38|.96|.96|.96|.56|.64|.35|.53|.31|.60|
|しゃ||||||||||||||||||||||0|.29|.23|.91|.66|.89|.75|.92|.05|.35|.29|.96|.76|.95|.83|.97|.97|.96|.98|.95|.74|.93|.81|.96|.19|.48|.42|.94|.72|.92|.79|.94|.95|
|しゅ|||||||||||||||||||||||0|.08|.91|.66|.89|.75|.92|.35|.05|.14|.96|.76|.95|.83|.97|.97|.96|.98|.95|.74|.93|.81|.96|.48|.19|.27|.94|.72|.92|.79|.94|.95|
|しょ||||||||||||||||||||||||0|.91|.66|.89|.75|.92|.29|.14|.05|.96|.76|.95|.83|.97|.97|.96|.98|.95|.74|.93|.81|.96|.42|.27|.19|.94|.72|.92|.79|.94|.95|
|た|||||||||||||||||||||||||0|.55|.59|.35|.46|.91|.91|.91|.40|.67|.71|.53|.61|.45|.74|.65|.41|.68|.72|.54|.62|.96|.96|.96|.25|.60|.64|.43|.53|.33|
|ち||||||||||||||||||||||||||0|.53|.20|.56|.65|.65|.65|.67|.40|.67|.45|.68|.71|.70|.72|.68|.41|.67|.46|.69|.75|.75|.75|.60|.25|.59|.32|.61|.64|
|つ|||||||||||||||||||||||||||0|.46|.16|.89|.89|.89|.71|.67|.40|.60|.43|.74|.45|.48|.72|.67|.41|.61|.44|.94|.94|.94|.64|.59|.25|.52|.30|.68|
|て||||||||||||||||||||||||||||0|.43|.74|.74|.74|.53|.45|.60|.40|.58|.57|.64|.62|.54|.46|.61|.41|.59|.82|.82|.82|.43|.32|.52|.25|.49|.48|
|と|||||||||||||||||||||||||||||0|.92|.92|.92|.61|.68|.43|.58|.40|.65|.48|.45|.62|.69|.44|.59|.41|.96|.96|.96|.53|.61|.30|.49|.25|.57|
|ちゃ||||||||||||||||||||||||||||||0|.29|.23|.95|.75|.94|.81|.96|.97|.95|.97|.96|.75|.94|.82|.96|.20|.50|.43|.93|.69|.91|.78|.93|.94|
|ちゅ|||||||||||||||||||||||||||||||0|.08|.95|.75|.94|.81|.96|.97|.95|.97|.96|.75|.94|.82|.96|.50|.20|.29|.93|.69|.91|.78|.93|.94|
|ちょ||||||||||||||||||||||||||||||||0|.95|.75|.94|.81|.96|.97|.95|.97|.96|.75|.94|.82|.96|.43|.29|.20|.93|.69|.91|.78|.93|.94|
|な|||||||||||||||||||||||||||||||||0|.55|.59|.35|.46|.22|.63|.51|.35|.65|.69|.50|.58|.94|.94|.94|.37|.66|.70|.51|.59|.43|
|に||||||||||||||||||||||||||||||||||0|.53|.20|.56|.59|.58|.60|.65|.35|.64|.41|.66|.73|.73|.73|.66|.37|.65|.43|.67|.70|
|ぬ|||||||||||||||||||||||||||||||||||0|.46|.16|.63|.22|.27|.69|.64|.35|.58|.39|.93|.93|.93|.70|.65|.37|.59|.41|.73|
|ね||||||||||||||||||||||||||||||||||||0|.43|.41|.51|.48|.50|.41|.58|.35|.55|.80|.80|.80|.51|.43|.59|.37|.57|.56|
|の|||||||||||||||||||||||||||||||||||||0|.51|.27|.22|.58|.66|.39|.55|.35|.95|.95|.95|.59|.67|.41|.57|.37|.63|
|にゃ||||||||||||||||||||||||||||||||||||||0|.59|.46|.42|.69|.72|.54|.62|.96|.96|.96|.43|.70|.73|.56|.63|.37|
|にゅ|||||||||||||||||||||||||||||||||||||||0|.16|.72|.68|.42|.62|.45|.94|.94|.94|.73|.69|.43|.63|.46|.70|
|にょ||||||||||||||||||||||||||||||||||||||||0|.62|.70|.45|.60|.42|.96|.96|.96|.63|.71|.46|.61|.43|.59|
|は|||||||||||||||||||||||||||||||||||||||||0|.55|.59|.35|.46|.91|.91|.91|.56|.78|.81|.66|.72|.60|
|ひ||||||||||||||||||||||||||||||||||||||||||0|.53|.20|.56|.65|.65|.65|.78|.56|.77|.59|.79|.81|
|ふ|||||||||||||||||||||||||||||||||||||||||||0|.46|.16|.89|.89|.89|.81|.77|.56|.72|.58|.84|
|へ||||||||||||||||||||||||||||||||||||||||||||0|.43|.74|.74|.74|.66|.59|.72|.56|.70|.69|
|ほ|||||||||||||||||||||||||||||||||||||||||||||0|.92|.92|.92|.72|.79|.58|.70|.56|.76|
|ひゃ||||||||||||||||||||||||||||||||||||||||||||||0|.29|.23|.99|.82|.98|.87|.99|.99|
|ひゅ|||||||||||||||||||||||||||||||||||||||||||||||0|.08|.99|.82|.98|.87|.99|.99|
|ひょ||||||||||||||||||||||||||||||||||||||||||||||||0|.99|.82|.98|.87|.99|.99|
|ま|||||||||||||||||||||||||||||||||||||||||||||||||0|.55|.59|.35|.46|.22|
|み||||||||||||||||||||||||||||||||||||||||||||||||||0|.53|.20|.56|.59|
|む|||||||||||||||||||||||||||||||||||||||||||||||||||0|.46|.16|.63|
|め||||||||||||||||||||||||||||||||||||||||||||||||||||0|.43|.41|
|も|||||||||||||||||||||||||||||||||||||||||||||||||||||0|.51|
|みゃ||||||||||||||||||||||||||||||||||||||||||||||||||||||0|
|みゅ|||||||||||||||||||||||||||||||||||||||||||||||||||||||
|みょ|||||||||||||||||||||||||||||||||||||||||||||||||||||||
|ら|||||||||||||||||||||||||||||||||||||||||||||||||||||||
|り|||||||||||||||||||||||||||||||||||||||||||||||||||||||
|る|||||||||||||||||||||||||||||||||||||||||||||||||||||||
|れ|||||||||||||||||||||||||||||||||||||||||||||||||||||||
|ろ|||||||||||||||||||||||||||||||||||||||||||||||||||||||
|りゃ|||||||||||||||||||||||||||||||||||||||||||||||||||||||
|りゅ|||||||||||||||||||||||||||||||||||||||||||||||||||||||
|りょ|||||||||||||||||||||||||||||||||||||||||||||||||||||||
|わ|||||||||||||||||||||||||||||||||||||||||||||||||||||||
|ゐ|||||||||||||||||||||||||||||||||||||||||||||||||||||||
|ゑ|||||||||||||||||||||||||||||||||||||||||||||||||||||||
|を|||||||||||||||||||||||||||||||||||||||||||||||||||||||
|ん|||||||||||||||||||||||||||||||||||||||||||||||||||||||
|が|||||||||||||||||||||||||||||||||||||||||||||||||||||||
|ぎ|||||||||||||||||||||||||||||||||||||||||||||||||||||||
|ぐ|||||||||||||||||||||||||||||||||||||||||||||||||||||||
|げ|||||||||||||||||||||||||||||||||||||||||||||||||||||||
|ご|||||||||||||||||||||||||||||||||||||||||||||||||||||||
|ぎゃ|||||||||||||||||||||||||||||||||||||||||||||||||||||||
|ぎゅ|||||||||||||||||||||||||||||||||||||||||||||||||||||||
|ぎょ|||||||||||||||||||||||||||||||||||||||||||||||||||||||
|ざ|||||||||||||||||||||||||||||||||||||||||||||||||||||||
|じ|||||||||||||||||||||||||||||||||||||||||||||||||||||||
|ず|||||||||||||||||||||||||||||||||||||||||||||||||||||||
|ぜ|||||||||||||||||||||||||||||||||||||||||||||||||||||||
|ぞ|||||||||||||||||||||||||||||||||||||||||||||||||||||||
|じゃ|||||||||||||||||||||||||||||||||||||||||||||||||||||||
|じゅ|||||||||||||||||||||||||||||||||||||||||||||||||||||||
|じょ|||||||||||||||||||||||||||||||||||||||||||||||||||||||
|だ|||||||||||||||||||||||||||||||||||||||||||||||||||||||
|ぢ|||||||||||||||||||||||||||||||||||||||||||||||||||||||
|づ|||||||||||||||||||||||||||||||||||||||||||||||||||||||
|で|||||||||||||||||||||||||||||||||||||||||||||||||||||||
|ど|||||||||||||||||||||||||||||||||||||||||||||||||||||||
|ぢゃ|||||||||||||||||||||||||||||||||||||||||||||||||||||||
|ぢゅ|||||||||||||||||||||||||||||||||||||||||||||||||||||||
|ぢょ|||||||||||||||||||||||||||||||||||||||||||||||||||||||
|ば|||||||||||||||||||||||||||||||||||||||||||||||||||||||
|び|||||||||||||||||||||||||||||||||||||||||||||||||||||||
|ぶ|||||||||||||||||||||||||||||||||||||||||||||||||||||||
|べ|||||||||||||||||||||||||||||||||||||||||||||||||||||||
|ぼ|||||||||||||||||||||||||||||||||||||||||||||||||||||||
|びゃ|||||||||||||||||||||||||||||||||||||||||||||||||||||||
|びゅ|||||||||||||||||||||||||||||||||||||||||||||||||||||||
|びょ|||||||||||||||||||||||||||||||||||||||||||||||||||||||
|ぱ|||||||||||||||||||||||||||||||||||||||||||||||||||||||
|ぴ|||||||||||||||||||||||||||||||||||||||||||||||||||||||
|ぷ|||||||||||||||||||||||||||||||||||||||||||||||||||||||
|ぺ|||||||||||||||||||||||||||||||||||||||||||||||||||||||
|ぽ|||||||||||||||||||||||||||||||||||||||||||||||||||||||
|ぴゃ|||||||||||||||||||||||||||||||||||||||||||||||||||||||
|ぴゅ|||||||||||||||||||||||||||||||||||||||||||||||||||||||
|ぴょ|||||||||||||||||||||||||||||||||||||||||||||||||||||||

**Fig. 6.** (On two pages) Values of Japanese phonetic distance for all syllable combinations

| みゅ | みょ | ら | り | る | れ | ろ | りゃ | りゅ | りょ | わ | ゐ | ゑ | を | ん | が | ぎ | ぐ | げ | ご | ぎゃ | ぎゅ | ぎょ | ざ | じ | ず | ぜ | ぞ | じゃ | じゅ | じょ | だ | ぢ | づ | で | ど | ぢゃ | ぢゅ | ぢょ | ば | び | ぶ | べ | ぼ | びゃ | びゅ | びょ | ぱ | ぴ | ぷ | ぺ | ぽ | ぴゃ | ぴゅ | ぴょ | |
|---|---|---|---|---|---|---|---|---|---|---|---|---|---|---|---|---|---|---|---|---|---|---|---|---|---|---|---|---|---|---|---|---|---|---|---|---|---|---|---|---|---|---|---|---|---|---|---|---|---|---|---|---|---|---|---|
| .68 | .57 | .64 | .84 | .87 | .73 | .79 | .67 | .89 | .82 | .61 | .82 | .71 | .77 | .95 | .48 | .73 | .76 | .60 | .67 | .53 | .79 | .70 | .41 | .68 | .72 | .54 | .62 | .96 | .96 | .96 | .33 | .64 | .68 | .48 | .57 | .94 | .94 | .94 | .31 | .63 | .67 | .47 | .56 | .94 | .94 | .94 | .23 | .59 | .63 | .41 | .51 | .92 | .92 | .92 | あ |
| .63 | .65 | .84 | .64 | .83 | .67 | .85 | .87 | .86 | .88 | .82 | .61 | .65 | .83 | .58 | .73 | .48 | .72 | .53 | .74 | .76 | .75 | .77 | .68 | .41 | .68 | .46 | .69 | .75 | .75 | .75 | .64 | .33 | .63 | .39 | .65 | .72 | .72 | .72 | .63 | .31 | .62 | .37 | .64 | .72 | .72 | .72 | .59 | .23 | .58 | .30 | .60 | .69 | .69 | .69 | い |
| .33 | .37 | .87 | .83 | .64 | .78 | .66 | .89 | .67 | .69 | .85 | .81 | .77 | .64 | .93 | .76 | .72 | .48 | .67 | .51 | .79 | .53 | .56 | .72 | .68 | .41 | .62 | .44 | .94 | .94 | .94 | .68 | .63 | .33 | .56 | .37 | .92 | .92 | .92 | .67 | .62 | .31 | .55 | .35 | .92 | .92 | .92 | .63 | .58 | .23 | .51 | .28 | .91 | .91 | .91 | う |
| .56 | .54 | .73 | .67 | .78 | .64 | .77 | .76 | .81 | .80 | .71 | .65 | .61 | .75 | .70 | .60 | .53 | .67 | .48 | .65 | .64 | .70 | .68 | .54 | .46 | .62 | .41 | .59 | .82 | .82 | .82 | .48 | .39 | .56 | .33 | .54 | .80 | .80 | .80 | .47 | .37 | .55 | .31 | .53 | .79 | .79 | .79 | .41 | .30 | .51 | .23 | .48 | .77 | .77 | .77 | え |
| .37 | .33 | .79 | .85 | .66 | .77 | .64 | .82 | .69 | .67 | .77 | .83 | .75 | .61 | .97 | .67 | .74 | .51 | .65 | .48 | .70 | .56 | .53 | .62 | .69 | .44 | .59 | .41 | .96 | .96 | .96 | .57 | .65 | .37 | .54 | .33 | .95 | .95 | .95 | .56 | .64 | .35 | .53 | .31 | .94 | .94 | .94 | .51 | .60 | .28 | .48 | .23 | .93 | .93 | .93 | お |
| .75 | .66 | .13 | .56 | .60 | .37 | .48 | .25 | .64 | .53 | .06 | .55 | .35 | .47 | .89 | .23 | .59 | .63 | .41 | .51 | .31 | .67 | .56 | .22 | .59 | .63 | .41 | .51 | .92 | .92 | .92 | .29 | .62 | .65 | .45 | .54 | .93 | .93 | .93 | .33 | .64 | .67 | .48 | .57 | .94 | .94 | .94 | .40 | .67 | .71 | .53 | .61 | .95 | .95 | .95 | や |
| .47 | .49 | .60 | .55 | .13 | .47 | .21 | .64 | .25 | .30 | .59 | .54 | .46 | .17 | .86 | .63 | .58 | .23 | .51 | .28 | .67 | .31 | .35 | .63 | .58 | .22 | .50 | .27 | .91 | .91 | .91 | .65 | .61 | .29 | .54 | .33 | .92 | .92 | .92 | .67 | .63 | .33 | .56 | .37 | .92 | .92 | .92 | .71 | .67 | .40 | .60 | .43 | .94 | .94 | .94 | ゆ |
| .49 | .47 | .48 | .57 | .21 | .44 | .13 | .53 | .30 | .25 | .47 | .56 | .43 | .06 | .90 | .51 | .60 | .28 | .48 | .23 | .56 | .35 | .31 | .51 | .60 | .27 | .48 | .22 | .93 | .93 | .93 | .54 | .63 | .33 | .51 | .29 | .94 | .94 | .94 | .57 | .65 | .37 | .54 | .33 | .95 | .95 | .95 | .61 | .69 | .43 | .58 | .40 | .96 | .96 | .96 | よ |
| .74 | .65 | .41 | .68 | .72 | .54 | .62 | .47 | .75 | .66 | .31 | .63 | .46 | .56 | .85 | .22 | .59 | .63 | .41 | .51 | .31 | .66 | .56 | .35 | .65 | .68 | .49 | .58 | .94 | .94 | .94 | .33 | .64 | .67 | .48 | .57 | .94 | .94 | .94 | .38 | .67 | .70 | .52 | .60 | .95 | .95 | .95 | .31 | .63 | .67 | .47 | .56 | .94 | .94 | .94 | か |
| .70 | .72 | .68 | .41 | .68 | .46 | .69 | .72 | .71 | .73 | .63 | .31 | .37 | .64 | .39 | .59 | .22 | .58 | .30 | .60 | .63 | .62 | .64 | .65 | .35 | .64 | .40 | .66 | .73 | .73 | .73 | .64 | .33 | .63 | .39 | .65 | .72 | .72 | .72 | .67 | .38 | .66 | .43 | .68 | .74 | .74 | .74 | .63 | .31 | .62 | .37 | .64 | .71 | .71 | .71 | き |
| .45 | .48 | .72 | .68 | .41 | .62 | .44 | .75 | .47 | .49 | .66 | .62 | .55 | .35 | .82 | .63 | .58 | .22 | .51 | .27 | .66 | .31 | .35 | .68 | .64 | .35 | .57 | .38 | .93 | .93 | .93 | .67 | .63 | .33 | .56 | .37 | .92 | .92 | .92 | .70 | .66 | .38 | .59 | .41 | .93 | .93 | .93 | .67 | .62 | .31 | .55 | .35 | .92 | .92 | .92 | く |
| .64 | .62 | .54 | .46 | .62 | .41 | .59 | .58 | .65 | .63 | .46 | .37 | .31 | .53 | .55 | .41 | .30 | .51 | .22 | .48 | .46 | .55 | .53 | .49 | .40 | .57 | .35 | .55 | .80 | .80 | .80 | .48 | .39 | .56 | .33 | .54 | .80 | .80 | .80 | .52 | .43 | .59 | .38 | .57 | .81 | .81 | .81 | .47 | .37 | .55 | .31 | .53 | .79 | .79 | .79 | け |
| .48 | .45 | .62 | .69 | .44 | .59 | .41 | .66 | .49 | .47 | .56 | .64 | .53 | .31 | .87 | .51 | .60 | .27 | .48 | .22 | .56 | .35 | .31 | .58 | .66 | .38 | .55 | .35 | .95 | .95 | .95 | .57 | .65 | .37 | .54 | .33 | .95 | .95 | .95 | .60 | .68 | .41 | .57 | .38 | .96 | .96 | .96 | .56 | .64 | .35 | .53 | .31 | .94 | .94 | .94 | こ |
| .71 | .61 | .47 | .72 | .75 | .58 | .66 | .41 | .72 | .62 | .38 | .66 | .51 | .60 | .88 | .31 | .63 | .66 | .46 | .56 | .22 | .63 | .51 | .41 | .68 | .72 | .54 | .62 | .96 | .96 | .96 | .40 | .67 | .71 | .53 | .61 | .95 | .95 | .95 | .44 | .70 | .73 | .56 | .64 | .96 | .96 | .96 | .38 | .67 | .70 | .52 | .60 | .95 | .95 | .95 | きゃ |
| .40 | .43 | .75 | .71 | .47 | .65 | .49 | .72 | .41 | .44 | .70 | .65 | .59 | .41 | .85 | .66 | .62 | .31 | .55 | .35 | .63 | .22 | .27 | .72 | .67 | .41 | .61 | .44 | .94 | .94 | .94 | .71 | .67 | .40 | .60 | .43 | .94 | .94 | .94 | .73 | .69 | .44 | .63 | .47 | .95 | .95 | .95 | .70 | .66 | .38 | .59 | .41 | .93 | .93 | .93 | きゅ |
| .43 | .40 | .66 | .73 | .49 | .63 | .47 | .62 | .44 | .41 | .60 | .67 | .57 | .38 | .89 | .56 | .64 | .35 | .53 | .31 | .51 | .27 | .22 | .62 | .69 | .44 | .59 | .41 | .96 | .96 | .96 | .61 | .69 | .43 | .58 | .40 | .96 | .96 | .96 | .64 | .71 | .47 | .61 | .44 | .97 | .97 | .97 | .60 | .68 | .41 | .57 | .38 | .96 | .96 | .96 | きょ |
| .70 | .60 | .33 | .64 | .68 | .48 | .57 | .40 | .71 | .61 | .35 | .65 | .49 | .58 | .92 | .35 | .65 | .68 | .49 | .58 | .41 | .72 | .62 | .22 | .59 | .63 | .41 | .51 | .92 | .92 | .92 | .24 | .60 | .64 | .42 | .52 | .92 | .92 | .92 | .25 | .60 | .64 | .43 | .53 | .93 | .93 | .93 | .13 | .56 | .60 | .37 | .48 | .91 | .91 | .91 | さ |
| .66 | .68 | .64 | .33 | .63 | .39 | .65 | .68 | .67 | .69 | .65 | .35 | .40 | .66 | .53 | .65 | .35 | .64 | .40 | .66 | .68 | .67 | .69 | .59 | .22 | .58 | .30 | .60 | .68 | .68 | .68 | .60 | .24 | .59 | .32 | .61 | .69 | .69 | .69 | .60 | .25 | .59 | .32 | .61 | .69 | .69 | .69 | .56 | .13 | .55 | .24 | .57 | .66 | .66 | .66 | し |
| .38 | .42 | .68 | .63 | .33 | .56 | .37 | .71 | .40 | .43 | .68 | .64 | .57 | .38 | .90 | .68 | .64 | .35 | .57 | .38 | .72 | .41 | .44 | .63 | .58 | .22 | .51 | .27 | .91 | .91 | .91 | .64 | .59 | .24 | .52 | .29 | .91 | .91 | .91 | .64 | .59 | .25 | .52 | .30 | .91 | .91 | .91 | .60 | .55 | .13 | .47 | .21 | .90 | .90 | .90 | す |
| .60 | .57 | .48 | .39 | .56 | .33 | .54 | .53 | .61 | .58 | .49 | .40 | .35 | .55 | .66 | .49 | .40 | .57 | .35 | .55 | .54 | .61 | .59 | .41 | .30 | .51 | .22 | .48 | .77 | .77 | .77 | .42 | .32 | .52 | .24 | .49 | .77 | .77 | .77 | .43 | .32 | .52 | .25 | .49 | .78 | .78 | .78 | .37 | .24 | .47 | .13 | .44 | .75 | .75 | .75 | せ |
| .42 | .38 | .57 | .65 | .37 | .54 | .33 | .61 | .43 | .40 | .58 | .66 | .55 | .35 | .94 | .58 | .66 | .38 | .55 | .35 | .62 | .44 | .41 | .51 | .60 | .27 | .48 | .22 | .93 | .93 | .93 | .52 | .61 | .29 | .49 | .24 | .93 | .93 | .93 | .53 | .61 | .30 | .49 | .25 | .93 | .93 | .93 | .48 | .57 | .21 | .44 | .13 | .92 | .92 | .92 | そ |
| .93 | .96 | .94 | .72 | .92 | .80 | .95 | .95 | .94 | .96 | .94 | .73 | .80 | .95 | .72 | .94 | .73 | .93 | .80 | .95 | .96 | .94 | .96 | .92 | .68 | .91 | .77 | .93 | .11 | .40 | .34 | .92 | .69 | .91 | .77 | .93 | .12 | .42 | .35 | .93 | .69 | .91 | .78 | .93 | .13 | .42 | .36 | .91 | .66 | .90 | .75 | .92 | .06 | .36 | .29 | しゃ |
| .93 | .96 | .94 | .72 | .92 | .80 | .95 | .95 | .94 | .96 | .94 | .73 | .80 | .95 | .72 | .94 | .73 | .93 | .80 | .95 | .96 | .94 | .96 | .92 | .68 | .91 | .77 | .93 | .40 | .11 | .19 | .92 | .69 | .91 | .77 | .93 | .42 | .12 | .20 | .93 | .69 | .91 | .78 | .93 | .42 | .13 | .21 | .91 | .66 | .90 | .75 | .92 | .36 | .06 | .14 | しゅ |
| .93 | .96 | .94 | .72 | .92 | .80 | .95 | .95 | .94 | .96 | .94 | .73 | .80 | .95 | .72 | .94 | .73 | .93 | .80 | .95 | .96 | .94 | .96 | .92 | .68 | .91 | .77 | .93 | .34 | .19 | .11 | .92 | .69 | .91 | .77 | .93 | .35 | .20 | .12 | .93 | .69 | .91 | .78 | .93 | .36 | .21 | .13 | .91 | .66 | .90 | .75 | .92 | .29 | .14 | .06 | しょ |
| .68 | .57 | .41 | .68 | .72 | .54 | .62 | .47 | .75 | .66 | .40 | .67 | .53 | .61 | .90 | .33 | .64 | .67 | .48 | .57 | .40 | .71 | .61 | .24 | .60 | .64 | .42 | .52 | .92 | .92 | .92 | .22 | .59 | .63 | .41 | .51 | .92 | .92 | .92 | .23 | .59 | .63 | .41 | .51 | .92 | .92 | .92 | .06 | .55 | .59 | .35 | .47 | .91 | .91 | .91 | た |
| .63 | .65 | .68 | .41 | .68 | .46 | .69 | .72 | .71 | .73 | .67 | .40 | .45 | .69 | .49 | .64 | .33 | .63 | .39 | .65 | .67 | .67 | .69 | .60 | .24 | .59 | .32 | .61 | .69 | .69 | .69 | .59 | .22 | .58 | .30 | .60 | .68 | .68 | .68 | .59 | .23 | .58 | .30 | .60 | .69 | .69 | .69 | .55 | .06 | .54 | .21 | .56 | .65 | .65 | .65 | ち |
| .33 | .37 | .72 | .68 | .41 | .62 | .44 | .75 | .47 | .49 | .71 | .67 | .60 | .43 | .88 | .67 | .63 | .33 | .56 | .37 | .71 | .40 | .43 | .64 | .59 | .24 | .52 | .29 | .91 | .91 | .91 | .63 | .58 | .22 | .51 | .27 | .91 | .91 | .91 | .63 | .58 | .23 | .51 | .28 | .91 | .91 | .91 | .59 | .54 | .06 | .46 | .17 | .89 | .89 | .89 | つ |
| .56 | .54 | .54 | .46 | .62 | .41 | .59 | .58 | .65 | .63 | .53 | .45 | .40 | .58 | .63 | .48 | .39 | .56 | .33 | .54 | .53 | .60 | .58 | .42 | .32 | .52 | .24 | .49 | .77 | .77 | .77 | .41 | .30 | .51 | .22 | .48 | .77 | .77 | .77 | .41 | .30 | .51 | .23 | .48 | .77 | .77 | .77 | .35 | .21 | .46 | .06 | .43 | .75 | .75 | .75 | て |
| .37 | .33 | .62 | .69 | .44 | .59 | .41 | .66 | .49 | .47 | .61 | .69 | .58 | .40 | .92 | .57 | .65 | .37 | .54 | .33 | .61 | .43 | .40 | .52 | .61 | .29 | .49 | .24 | .93 | .93 | .93 | .51 | .60 | .27 | .48 | .22 | .93 | .93 | .93 | .51 | .60 | .28 | .48 | .23 | .93 | .93 | .93 | .47 | .56 | .17 | .43 | .06 | .92 | .92 | .92 | と |
| .92 | .95 | .96 | .75 | .94 | .82 | .96 | .97 | .95 | .98 | .95 | .75 | .82 | .96 | .70 | .94 | .72 | .92 | .80 | .95 | .95 | .94 | .96 | .92 | .69 | .91 | .77 | .93 | .12 | .42 | .35 | .92 | .68 | .91 | .77 | .93 | .11 | .40 | .34 | .92 | .69 | .91 | .77 | .93 | .11 | .41 | .34 | .91 | .65 | .89 | .75 | .92 | .03 | .32 | .26 | ちゃ |
| .92 | .95 | .96 | .75 | .94 | .82 | .96 | .97 | .95 | .98 | .95 | .75 | .82 | .96 | .70 | .94 | .72 | .92 | .80 | .95 | .95 | .94 | .96 | .92 | .69 | .91 | .77 | .93 | .42 | .12 | .20 | .92 | .68 | .91 | .77 | .93 | .40 | .11 | .19 | .92 | .69 | .91 | .77 | .93 | .41 | .11 | .19 | .91 | .65 | .89 | .75 | .92 | .32 | .03 | .11 | ちゅ |
| .92 | .95 | .96 | .75 | .94 | .82 | .96 | .97 | .95 | .98 | .95 | .75 | .82 | .96 | .70 | .94 | .72 | .92 | .80 | .95 | .95 | .94 | .96 | .92 | .69 | .91 | .77 | .93 | .35 | .20 | .12 | .92 | .68 | .91 | .77 | .93 | .34 | .19 | .11 | .92 | .69 | .91 | .77 | .93 | .34 | .19 | .11 | .91 | .65 | .89 | .75 | .92 | .26 | .11 | .03 | ちょ |
| .73 | .63 | .47 | .72 | .76 | .59 | .66 | .52 | .79 | .70 | .33 | .64 | .48 | .57 | .81 | .13 | .56 | .60 | .37 | .48 | .25 | .64 | .53 | .38 | .67 | .70 | .52 | .60 | .95 | .95 | .95 | .33 | .64 | .67 | .48 | .57 | .94 | .94 | .94 | .39 | .67 | .71 | .52 | .60 | .95 | .95 | .95 | .45 | .71 | .74 | .57 | .64 | .96 | .96 | .96 | な |
| .69 | .71 | .72 | .47 | .71 | .52 | .73 | .76 | .75 | .77 | .64 | .33 | .39 | .65 | .29 | .56 | .13 | .55 | .24 | .57 | .60 | .59 | .61 | .67 | .38 | .66 | .43 | .68 | .74 | .74 | .74 | .64 | .33 | .63 | .39 | .65 | .72 | .72 | .72 | .67 | .39 | .66 | .44 | .68 | .74 | .74 | .74 | .71 | .45 | .70 | .49 | .72 | .77 | .77 | .77 | に |
| .43 | .46 | .76 | .71 | .47 | .66 | .50 | .79 | .52 | .55 | .68 | .63 | .56 | .37 | .78 | .60 | .55 | .13 | .47 | .21 | .64 | .25 | .30 | .70 | .66 | .38 | .59 | .41 | .93 | .93 | .93 | .67 | .63 | .33 | .56 | .37 | .92 | .92 | .92 | .71 | .66 | .39 | .60 | .42 | .94 | .94 | .94 | .74 | .70 | .45 | .64 | .48 | .95 | .95 | .95 | ぬ |
| .63 | .61 | .59 | .52 | .66 | .47 | .64 | .63 | .69 | .67 | .48 | .39 | .33 | .54 | .49 | .37 | .24 | .47 | .13 | .44 | .43 | .52 | .49 | .52 | .43 | .59 | .38 | .57 | .81 | .81 | .81 | .48 | .39 | .56 | .33 | .54 | .80 | .80 | .80 | .52 | .44 | .60 | .39 | .58 | .81 | .81 | .81 | .57 | .49 | .64 | .45 | .62 | .83 | .83 | .83 | ね |
| .46 | .43 | .66 | .73 | .50 | .64 | .47 | .70 | .55 | .52 | .57 | .65 | .54 | .33 | .83 | .48 | .57 | .21 | .44 | .13 | .53 | .30 | .25 | .60 | .68 | .41 | .57 | .38 | .96 | .96 | .96 | .57 | .65 | .37 | .54 | .33 | .95 | .95 | .95 | .60 | .68 | .42 | .58 | .39 | .96 | .96 | .96 | .64 | .72 | .48 | .62 | .45 | .97 | .97 | .97 | の |
| .70 | .59 | .52 | .76 | .79 | .63 | .70 | .47 | .76 | .66 | .40 | .68 | .53 | .61 | .84 | .25 | .60 | .64 | .43 | .53 | .13 | .60 | .48 | .44 | .70 | .73 | .56 | .64 | .96 | .96 | .96 | .40 | .67 | .71 | .53 | .61 | .95 | .95 | .95 | .45 | .71 | .74 | .57 | .64 | .96 | .96 | .96 | .50 | .74 | .77 | .61 | .68 | .98 | .98 | .98 | にゃ |
| .37 | .41 | .79 | .75 | .52 | .69 | .55 | .76 | .47 | .50 | .71 | .67 | .61 | .43 | .81 | .64 | .59 | .25 | .52 | .30 | .60 | .13 | .21 | .73 | .69 | .44 | .63 | .47 | .95 | .95 | .95 | .71 | .67 | .40 | .60 | .43 | .94 | .94 | .94 | .74 | .70 | .45 | .64 | .48 | .95 | .95 | .95 | .77 | .73 | .50 | .67 | .52 | .96 | .96 | .96 | にゅ |
| .41 | .37 | .70 | .77 | .55 | .67 | .52 | .66 | .50 | .47 | .61 | .69 | .58 | .40 | .86 | .53 | .61 | .30 | .49 | .25 | .48 | .21 | .13 | .64 | .71 | .47 | .61 | .44 | .97 | .97 | .97 | .61 | .68 | .43 | .58 | .40 | .96 | .96 | .96 | .64 | .72 | .48 | .62 | .45 | .97 | .97 | .97 | .68 | .75 | .52 | .65 | .50 | .98 | .98 | .98 | にょ |
| .84 | .76 | .37 | .66 | .69 | .51 | .59 | .43 | .73 | .63 | .26 | .60 | .43 | .53 | .89 | .30 | .62 | .66 | .46 | .55 | .37 | .69 | .59 | .44 | .70 | .73 | .56 | .64 | .96 | .96 | .96 | .46 | .72 | .75 | .58 | .65 | .97 | .97 | .97 | .51 | .75 | .78 | .62 | .69 | .98 | .98 | .98 | .47 | .72 | .75 | .58 | .66 | .97 | .97 | .97 | は |
| .80 | .82 | .66 | .37 | .65 | .42 | .67 | .69 | .69 | .70 | .60 | .26 | .33 | .62 | .46 | .62 | .30 | .61 | .36 | .63 | .66 | .65 | .67 | .70 | .44 | .69 | .48 | .71 | .76 | .76 | .76 | .72 | .46 | .71 | .51 | .73 | .77 | .77 | .77 | .75 | .51 | .74 | .55 | .76 | .80 | .80 | .80 | .72 | .47 | .71 | .51 | .73 | .78 | .78 | .78 | ひ |
| .60 | .62 | .69 | .65 | .37 | .59 | .40 | .73 | .43 | .46 | .64 | .59 | .52 | .30 | .86 | .66 | .61 | .30 | .55 | .34 | .69 | .37 | .41 | .73 | .69 | .44 | .63 | .47 | .95 | .95 | .95 | .75 | .71 | .46 | .65 | .49 | .95 | .95 | .95 | .78 | .74 | .51 | .69 | .54 | .97 | .97 | .97 | .75 | .71 | .47 | .65 | .49 | .95 | .95 | .95 | ふ |
| .75 | .74 | .51 | .42 | .59 | .37 | .56 | .55 | .63 | .60 | .43 | .33 | .26 | .50 | .60 | .46 | .36 | .55 | .30 | .52 | .51 | .59 | .56 | .56 | .48 | .63 | .44 | .61 | .83 | .83 | .83 | .58 | .51 | .65 | .46 | .63 | .84 | .84 | .84 | .62 | .55 | .69 | .51 | .67 | .86 | .86 | .86 | .58 | .51 | .65 | .47 | .63 | .84 | .84 | .84 | へ |
| .62 | .60 | .59 | .67 | .40 | .56 | .37 | .63 | .46 | .43 | .53 | .62 | .50 | .26 | .90 | .55 | .63 | .34 | .52 | .30 | .59 | .41 | .37 | .64 | .71 | .47 | .61 | .44 | .97 | .97 | .97 | .65 | .73 | .49 | .63 | .46 | .98 | .98 | .98 | .69 | .76 | .54 | .67 | .51 | .99 | .99 | .99 | .66 | .73 | .49 | .63 | .47 | .98 | .98 | .98 | ほ |
| .99 | .99 | .95 | .74 | .93 | .81 | .95 | .96 | .95 | .97 | .93 | .70 | .78 | .93 | .68 | .93 | .71 | .92 | .79 | .94 | .95 | .93 | .95 | .96 | .76 | .95 | .83 | .97 | .22 | .51 | .45 | .97 | .77 | .95 | .84 | .98 | .23 | .53 | .46 | .98 | .80 | .97 | .86 | .99 | .26 | .55 | .49 | .97 | .78 | .95 | .84 | .98 | .23 | .53 | .46 | ひゃ |
| .99 | .99 | .95 | .74 | .93 | .81 | .95 | .96 | .95 | .97 | .93 | .70 | .78 | .93 | .68 | .93 | .71 | .92 | .79 | .94 | .95 | .93 | .95 | .96 | .76 | .95 | .83 | .97 | .51 | .22 | .30 | .97 | .77 | .95 | .84 | .98 | .53 | .23 | .31 | .98 | .80 | .97 | .86 | .99 | .55 | .26 | .34 | .97 | .78 | .95 | .84 | .98 | .53 | .23 | .31 | ひゅ |
| .99 | .99 | .95 | .74 | .93 | .81 | .95 | .96 | .95 | .97 | .93 | .70 | .78 | .93 | .68 | .93 | .71 | .92 | .79 | .94 | .95 | .93 | .95 | .96 | .76 | .95 | .83 | .97 | .45 | .30 | .22 | .97 | .77 | .95 | .84 | .98 | .46 | .31 | .23 | .98 | .80 | .97 | .86 | .99 | .49 | .34 | .26 | .97 | .78 | .95 | .84 | .98 | .46 | .31 | .23 | ひょ |
| .63 | .51 | .47 | .72 | .76 | .59 | .66 | .52 | .79 | .70 | .45 | .71 | .57 | .65 | .90 | .33 | .64 | .67 | .48 | .57 | .40 | .71 | .61 | .23 | .59 | .63 | .41 | .51 | .92 | .92 | .92 | .13 | .56 | .60 | .37 | .48 | .91 | .91 | .91 | .11 | .56 | .60 | .36 | .47 | .91 | .91 | .91 | .24 | .60 | .64 | .42 | .52 | .92 | .92 | .92 | ま |
| .58 | .60 | .72 | .47 | .71 | .52 | .73 | .76 | .75 | .77 | .71 | .45 | .50 | .72 | .48 | .64 | .33 | .63 | .39 | .65 | .67 | .67 | .68 | .59 | .23 | .58 | .30 | .60 | .69 | .69 | .69 | .56 | .13 | .55 | .24 | .57 | .66 | .66 | .66 | .56 | .11 | .55 | .23 | .57 | .66 | .66 | .66 | .60 | .24 | .59 | .32 | .61 | .69 | .69 | .69 | み |
| .22 | .27 | .76 | .71 | .47 | .66 | .50 | .79 | .52 | .55 | .74 | .70 | .64 | .48 | .87 | .67 | .63 | .33 | .56 | .37 | .71 | .40 | .43 | .63 | .58 | .23 | .51 | .28 | .91 | .91 | .91 | .60 | .55 | .13 | .47 | .21 | .90 | .90 | .90 | .60 | .55 | .11 | .47 | .20 | .89 | .89 | .89 | .64 | .59 | .24 | .52 | .29 | .91 | .91 | .91 | む |
| .51 | .48 | .59 | .52 | .66 | .47 | .64 | .63 | .69 | .67 | .57 | .50 | .45 | .62 | .62 | .48 | .39 | .56 | .33 | .54 | .53 | .60 | .58 | .41 | .30 | .51 | .23 | .48 | .77 | .77 | .77 | .37 | .24 | .47 | .13 | .44 | .75 | .75 | .75 | .36 | .23 | .47 | .11 | .44 | .75 | .75 | .75 | .42 | .32 | .52 | .24 | .49 | .77 | .77 | .77 | め |
| .27 | .22 | .66 | .73 | .50 | .64 | .47 | .70 | .55 | .52 | .65 | .72 | .62 | .45 | .91 | .57 | .65 | .37 | .54 | .33 | .61 | .43 | .40 | .51 | .60 | .28 | .48 | .23 | .93 | .93 | .93 | .48 | .57 | .21 | .44 | .13 | .92 | .92 | .92 | .47 | .57 | .20 | .44 | .11 | .92 | .92 | .92 | .52 | .61 | .29 | .49 | .24 | .93 | .93 | .93 | も |
| .59 | .46 | .52 | .76 | .79 | .63 | .70 | .47 | .76 | .66 | .50 | .74 | .61 | .68 | .92 | .40 | .67 | .71 | .53 | .61 | .33 | .67 | .57 | .31 | .63 | .67 | .47 | .56 | .94 | .94 | .94 | .25 | .60 | .64 | .43 | .53 | .93 | .93 | .93 | .24 | .60 | .64 | .42 | .52 | .92 | .92 | .92 | .33 | .64 | .67 | .48 | .57 | .94 | .94 | .94 | みゃ |
| 0 | .16 | .79 | .75 | .52 | .69 | .55 | .76 | .47 | .50 | .77 | .73 | .68 | .53 | .89 | .71 | .67 | .40 | .60 | .43 | .67 | .33 | .37 | .67 | .62 | .31 | .55 | .35 | .92 | .92 | .92 | .64 | .59 | .25 | .52 | .30 | .91 | .91 | .91 | .64 | .59 | .24 | .52 | .29 | .91 | .91 | .91 | .67 | .63 | .33 | .56 | .37 | .92 | .92 | .92 | みゅ |
| | 0 | .70 | .77 | .55 | .67 | .52 | .66 | .50 | .47 | .68 | .75 | .66 | .50 | .94 | .61 | .68 | .43 | .58 | .40 | .57 | .37 | .33 | .56 | .64 | .35 | .53 | .31 | .94 | .94 | .94 | .53 | .61 | .30 | .49 | .25 | .93 | .93 | .93 | .52 | .61 | .29 | .49 | .24 | .93 | .93 | .93 | .57 | .65 | .37 | .54 | .33 | .95 | .95 | .95 | みょ |
| | | 0 | .55 | .59 | .35 | .46 | .22 | .63 | .51 | .17 | .57 | .38 | .49 | .94 | .35 | .65 | .68 | .49 | .58 | .41 | .72 | .62 | .25 | .60 | .64 | .43 | .53 | .93 | .93 | .93 | .35 | .65 | .68 | .49 | .58 | .94 | .94 | .94 | .38 | .66 | .70 | .51 | .60 | .95 | .95 | .95 | .44 | .70 | .73 | .56 | .63 | .96 | .96 | .96 | ら |
| | | | 0 | .53 | .20 | .56 | .59 | .58 | .60 | .57 | .17 | .26 | .58 | .56 | .65 | .35 | .64 | .41 | .66 | .68 | .68 | .69 | .60 | .25 | .59 | .32 | .61 | .69 | .69 | .69 | .65 | .35 | .64 | .41 | .66 | .73 | .73 | .73 | .66 | .38 | .65 | .43 | .67 | .74 | .74 | .74 | .70 | .44 | .69 | .48 | .71 | .76 | .76 | .76 | り |
| | | | | 0 | .46 | .16 | .63 | .22 | .27 | .61 | .56 | .49 | .23 | .91 | .68 | .64 | .35 | .58 | .39 | .72 | .41 | .44 | .64 | .59 | .25 | .52 | .30 | .91 | .91 | .91 | .68 | .64 | .35 | .58 | .39 | .93 | .93 | .93 | .70 | .65 | .38 | .59 | .41 | .93 | .93 | .93 | .73 | .69 | .44 | .63 | .46 | .95 | .95 | .95 | る |
| | | | | | 0 | .43 | .41 | .51 | .48 | .38 | .26 | .17 | .46 | .68 | .49 | .41 | .58 | .35 | .55 | .54 | .62 | .59 | .43 | .32 | .52 | .25 | .49 | .78 | .78 | .78 | .49 | .41 | .58 | .35 | .55 | .80 | .80 | .80 | .51 | .43 | .59 | .38 | .57 | .81 | .81 | .81 | .56 | .48 | .63 | .44 | .61 | .83 | .83 | .83 | れ |
| | | | | | | 0 | .51 | .27 | .22 | .49 | .58 | .46 | .17 | .96 | .58 | .66 | .39 | .55 | .35 | .62 | .44 | .41 | .53 | .61 | .30 | .49 | .25 | .93 | .93 | .93 | .58 | .66 | .39 | .55 | .35 | .95 | .95 | .95 | .60 | .67 | .41 | .57 | .38 | .96 | .96 | .96 | .63 | .71 | .46 | .61 | .44 | .97 | .97 | .97 | ろ |
| | | | | | | | 0 | .59 | .46 | .27 | .61 | .44 | .54 | .97 | .41 | .68 | .72 | .54 | .62 | .35 | .68 | .58 | .33 | .64 | .68 | .48 | .57 | .94 | .94 | .94 | .41 | .68 | .72 | .54 | .62 | .96 | .96 | .96 | .44 | .70 | .73 | .56 | .63 | .96 | .96 | .96 | .49 | .73 | .76 | .60 | .67 | .97 | .97 | .97 | りゃ |
| | | | | | | | | 0 | .16 | .65 | .60 | .53 | .32 | .94 | .72 | .68 | .41 | .62 | .44 | .68 | .35 | .39 | .68 | .63 | .33 | .56 | .37 | .92 | .92 | .92 | .72 | .68 | .41 | .62 | .44 | .94 | .94 | .94 | .73 | .69 | .44 | .63 | .46 | .95 | .95 | .95 | .76 | .72 | .49 | .67 | .51 | .96 | .96 | .96 | りゅ |
| | | | | | | | | | 0 | .54 | .62 | .51 | .27 | .98 | .62 | .69 | .44 | .59 | .41 | .58 | .39 | .35 | .57 | .65 | .37 | .54 | .33 | .95 | .95 | .95 | .62 | .69 | .44 | .59 | .41 | .96 | .96 | .96 | .63 | .71 | .46 | .61 | .44 | .97 | .97 | .97 | .67 | .74 | .51 | .65 | .49 | .98 | .98 | .98 | りょ |
| | | | | | | | | | | 0 | .55 | .35 | .46 | .88 | .22 | .59 | .63 | .41 | .51 | .31 | .66 | .56 | .27 | .61 | .65 | .44 | .54 | .93 | .93 | .93 | .33 | .64 | .67 | .48 | .57 | .94 | .94 | .94 | .38 | .67 | .70 | .52 | .60 | .95 | .95 | .95 | .44 | .70 | .73 | .56 | .64 | .96 | .96 | .96 | わ |
| | | | | | | | | | | | 0 | .20 | .56 | .44 | .59 | .22 | .58 | .30 | .60 | .63 | .62 | .64 | .61 | .27 | .60 | .34 | .62 | .70 | .70 | .70 | .64 | .33 | .63 | .39 | .65 | .72 | .72 | .72 | .67 | .38 | .66 | .43 | .68 | .74 | .74 | .74 | .70 | .44 | .69 | .48 | .71 | .76 | .76 | .76 | ゐ |
| | | | | | | | | | | | | 0 | .43 | .59 | .41 | .30 | .51 | .22 | .48 | .46 | .55 | .53 | .44 | .34 | .53 | .27 | .51 | .78 | .78 | .78 | .48 | .39 | .56 | .33 | .54 | .80 | .80 | .80 | .52 | .43 | .59 | .38 | .57 | .81 | .81 | .81 | .56 | .48 | .63 | .44 | .61 | .83 | .83 | .83 | ゑ |
| | | | | | | | | | | | | | 0 | .89 | .51 | .60 | .27 | .48 | .22 | .56 | .35 | .31 | .54 | .62 | .32 | .51 | .27 | .94 | .94 | .94 | .57 | .65 | .37 | .54 | .33 | .95 | .95 | .95 | .60 | .68 | .41 | .57 | .38 | .96 | .96 | .96 | .64 | .71 | .47 | .61 | .44 | .97 | .97 | .97 | を |
| | | | | | | | | | | | | | | 0 | .82 | .32 | .79 | .51 | .84 | .85 | .82 | .87 | .90 | .48 | .87 | .62 | .91 | .69 | .69 | .69 | .88 | .44 | .85 | .59 | .89 | .67 | .67 | .67 | .90 | .49 | .87 | .63 | .92 | .69 | .69 | .69 | .93 | .53 | .90 | .66 | .94 | .72 | .72 | .72 | ん |
| | | | | | | | | | | | | | | | 0 | .55 | .59 | .35 | .46 | .22 | .63 | .51 | .27 | .61 | .65 | .44 | .54 | .93 | .93 | .93 | .25 | .60 | .64 | .43 | .52 | .93 | .93 | .93 | .31 | .63 | .67 | .47 | .56 | .94 | .94 | .94 | .38 | .67 | .70 | .52 | .60 | .95 | .95 | .95 | が |
| | | | | | | | | | | | | | | | | 0 | .53 | .20 | .56 | .59 | .58 | .60 | .61 | .27 | .60 | .34 | .62 | .70 | .70 | .70 | .60 | .25 | .59 | .32 | .61 | .69 | .69 | .69 | .63 | .31 | .62 | .37 | .64 | .71 | .71 | .71 | .67 | .38 | .66 | .43 | .68 | .74 | .74 | .74 | ぎ |
| | | | | | | | | | | | | | | | | | 0 | .46 | .16 | .63 | .22 | .27 | .65 | .60 | .27 | .53 | .32 | .91 | .91 | .91 | .64 | .59 | .25 | .52 | .30 | .91 | .91 | .91 | .67 | .62 | .31 | .55 | .35 | .92 | .92 | .92 | .70 | .66 | .38 | .59 | .41 | .93 | .93 | .93 | ぐ |
| | | | | | | | | | | | | | | | | | | 0 | .43 | .41 | .51 | .48 | .44 | .34 | .53 | .27 | .51 | .78 | .78 | .78 | .43 | .32 | .52 | .25 | .49 | .77 | .77 | .77 | .47 | .37 | .55 | .31 | .53 | .79 | .79 | .79 | .52 | .43 | .59 | .38 | .57 | .81 | .81 | .81 | げ |
| | | | | | | | | | | | | | | | | | | | 0 | .51 | .27 | .22 | .54 | .62 | .32 | .51 | .27 | .94 | .94 | .94 | .52 | .61 | .30 | .49 | .25 | .93 | .93 | .93 | .56 | .64 | .35 | .53 | .31 | .94 | .94 | .94 | .60 | .68 | .41 | .57 | .38 | .96 | .96 | .96 | ご |
| | | | | | | | | | | | | | | | | | | | | 0 | .59 | .46 | .35 | .65 | .68 | .49 | .58 | .94 | .94 | .94 | .33 | .64 | .67 | .48 | .57 | .94 | .94 | .94 | .38 | .67 | .70 | .52 | .60 | .95 | .95 | .95 | .44 | .70 | .73 | .56 | .64 | .96 | .96 | .96 | ぎゃ |
| | | | | | | | | | | | | | | | | | | | | | 0 | .16 | .68 | .64 | .35 | .57 | .38 | .93 | .93 | .93 | .67 | .63 | .33 | .56 | .37 | .92 | .92 | .92 | .70 | .66 | .38 | .59 | .41 | .93 | .93 | .93 | .73 | .69 | .44 | .63 | .47 | .95 | .95 | .95 | ぎゅ |
| | | | | | | | | | | | | | | | | | | | | | | 0 | .58 | .66 | .38 | .55 | .35 | .95 | .95 | .95 | .57 | .65 | .37 | .54 | .33 | .95 | .95 | .95 | .60 | .68 | .41 | .57 | .38 | .96 | .96 | .96 | .64 | .71 | .47 | .61 | .44 | .97 | .97 | .97 | ぎょ |
| | | | | | | | | | | | | | | | | | | | | | | | 0 | .55 | .59 | .35 | .46 | .91 | .91 | .91 | .11 | .56 | .60 | .36 | .47 | .91 | .91 | .91 | .13 | .56 | .60 | .37 | .48 | .91 | .91 | .91 | .25 | .60 | .64 | .43 | .53 | .93 | .93 | .93 | ざ |
| | | | | | | | | | | | | | | | | | | | | | | | | 0 | .53 | .20 | .56 | .65 | .65 | .65 | .56 | .11 | .55 | .23 | .57 | .66 | .66 | .66 | .56 | .13 | .55 | .24 | .57 | .66 | .66 | .66 | .60 | .25 | .59 | .32 | .61 | .69 | .69 | .69 | じ |
| | | | | | | | | | | | | | | | | | | | | | | | | | 0 | .46 | .16 | .89 | .89 | .89 | .60 | .55 | .11 | .47 | .20 | .89 | .89 | .89 | .60 | .55 | .13 | .47 | .21 | .90 | .90 | .90 | .64 | .59 | .25 | .52 | .30 | .91 | .91 | .91 | ず |
| | | | | | | | | | | | | | | | | | | | | | | | | | | 0 | .43 | .74 | .74 | .74 | .36 | .23 | .47 | .11 | .44 | .75 | .75 | .75 | .37 | .24 | .47 | .13 | .44 | .75 | .75 | .75 | .43 | .32 | .52 | .25 | .49 | .78 | .78 | .78 | ぜ |
| | | | | | | | | | | | | | | | | | | | | | | | | | | | 0 | .92 | .92 | .92 | .47 | .57 | .20 | .44 | .11 | .92 | .92 | .92 | .48 | .57 | .21 | .44 | .13 | .92 | .92 | .92 | .53 | .61 | .30 | .49 | .25 | .93 | .93 | .93 | ぞ |
| | | | | | | | | | | | | | | | | | | | | | | | | | | | | 0 | .29 | .23 | .91 | .66 | .89 | .75 | .92 | .05 | .35 | .29 | .91 | .66 | .90 | .75 | .92 | .06 | .36 | .29 | .93 | .69 | .91 | .78 | .93 | .13 | .42 | .36 | じゃ |
| | | | | | | | | | | | | | | | | | | | | | | | | | | | | | 0 | .08 | .91 | .66 | .89 | .75 | .92 | .35 | .05 | .14 | .91 | .66 | .90 | .75 | .92 | .36 | .06 | .14 | .93 | .69 | .91 | .78 | .93 | .42 | .13 | .21 | じゅ |
| | | | | | | | | | | | | | | | | | | | | | | | | | | | | | | 0 | .91 | .66 | .89 | .75 | .92 | .29 | .14 | .05 | .91 | .66 | .90 | .75 | .92 | .29 | .14 | .06 | .93 | .69 | .91 | .78 | .93 | .36 | .21 | .13 | じょ |
| | | | | | | | | | | | | | | | | | | | | | | | | | | | | | | | 0 | .55 | .59 | .35 | .46 | .91 | .91 | .91 | .06 | .55 | .59 | .35 | .47 | .91 | .91 | .91 | .23 | .59 | .63 | .41 | .51 | .92 | .92 | .92 | だ |
| | | | | | | | | | | | | | | | | | | | | | | | | | | | | | | | | 0 | .53 | .20 | .56 | .65 | .65 | .65 | .55 | .06 | .54 | .21 | .56 | .65 | .65 | .65 | .59 | .23 | .58 | .30 | .60 | .69 | .69 | .69 | ぢ |
| | | | | | | | | | | | | | | | | | | | | | | | | | | | | | | | | | 0 | .46 | .16 | .89 | .89 | .89 | .59 | .54 | .06 | .46 | .17 | .89 | .89 | .89 | .63 | .58 | .23 | .51 | .28 | .91 | .91 | .91 | づ |
| | | | | | | | | | | | | | | | | | | | | | | | | | | | | | | | | | | 0 | .43 | .74 | .74 | .74 | .35 | .21 | .46 | .06 | .43 | .75 | .75 | .75 | .41 | .30 | .51 | .23 | .48 | .77 | .77 | .77 | で |
| | | | | | | | | | | | | | | | | | | | | | | | | | | | | | | | | | | | 0 | .92 | .92 | .92 | .47 | .56 | .17 | .43 | .06 | .92 | .92 | .92 | .51 | .60 | .28 | .48 | .23 | .93 | .93 | .93 | ど |
| | | | | | | | | | | | | | | | | | | | | | | | | | | | | | | | | | | | | 0 | .29 | .23 | .91 | .65 | .89 | .75 | .92 | .03 | .32 | .26 | .92 | .69 | .91 | .77 | .93 | .11 | .41 | .34 | ぢゃ |
| | | | | | | | | | | | | | | | | | | | | | | | | | | | | | | | | | | | | | 0 | .08 | .91 | .65 | .89 | .75 | .92 | .32 | .03 | .11 | .92 | .69 | .91 | .77 | .93 | .41 | .11 | .19 | ぢゅ |
| | | | | | | | | | | | | | | | | | | | | | | | | | | | | | | | | | | | | | | 0 | .91 | .65 | .89 | .75 | .92 | .26 | .11 | .03 | .92 | .69 | .91 | .77 | .93 | .34 | .19 | .11 | ぢょ |
| | | | | | | | | | | | | | | | | | | | | | | | | | | | | | | | | | | | | | | | 0 | .55 | .59 | .35 | .46 | .91 | .91 | .91 | .22 | .59 | .63 | .41 | .51 | .92 | .92 | .92 | ば |
| | | | | | | | | | | | | | | | | | | | | | | | | | | | | | | | | | | | | | | | | 0 | .53 | .20 | .56 | .65 | .65 | .65 | .59 | .22 | .58 | .30 | .60 | .68 | .68 | .68 | び |
| | | | | | | | | | | | | | | | | | | | | | | | | | | | | | | | | | | | | | | | | | 0 | .46 | .16 | .89 | .89 | .89 | .63 | .58 | .22 | .51 | .27 | .91 | .91 | .91 | ぶ |
| | | | | | | | | | | | | | | | | | | | | | | | | | | | | | | | | | | | | | | | | | | 0 | .43 | .74 | .74 | .74 | .41 | .30 | .51 | .22 | .48 | .77 | .77 | .77 | べ |
| | | | | | | | | | | | | | | | | | | | | | | | | | | | | | | | | | | | | | | | | | | | 0 | .92 | .92 | .92 | .51 | .60 | .27 | .48 | .22 | .93 | .93 | .93 | ぼ |
| | | | | | | | | | | | | | | | | | | | | | | | | | | | | | | | | | | | | | | | | | | | | 0 | .29 | .23 | .92 | .68 | .91 | .77 | .93 | .11 | .40 | .34 | びゃ |
| | | | | | | | | | | | | | | | | | | | | | | | | | | | | | | | | | | | | | | | | | | | | | 0 | .08 | .92 | .68 | .91 | .77 | .93 | .40 | .11 | .19 | びゅ |
| | | | | | | | | | | | | | | | | | | | | | | | | | | | | | | | | | | | | | | | | | | | | | | 0 | .92 | .68 | .91 | .77 | .93 | .34 | .19 | .11 | びょ |
| | | | | | | | | | | | | | | | | | | | | | | | | | | | | | | | | | | | | | | | | | | | | | | | 0 | .55 | .59 | .35 | .46 | .91 | .91 | .91 | ぱ |
| | | | | | | | | | | | | | | | | | | | | | | | | | | | | | | | | | | | | | | | | | | | | | | | | 0 | .53 | .20 | .56 | .65 | .65 | .65 | ぴ |
| | | | | | | | | | | | | | | | | | | | | | | | | | | | | | | | | | | | | | | | | | | | | | | | | | 0 | .46 | .16 | .89 | .89 | .89 | ぷ |
| | | | | | | | | | | | | | | | | | | | | | | | | | | | | | | | | | | | | | | | | | | | | | | | | | | 0 | .43 | .74 | .74 | .74 | ぺ |
| | | | | | | | | | | | | | | | | | | | | | | | | | | | | | | | | | | | | | | | | | | | | | | | | | | | 0 | .92 | .92 | .92 | ぽ |
| | | | | | | | | | | | | | | | | | | | | | | | | | | | | | | | | | | | | | | | | | | | | | | | | | | | | 0 | .29 | .23 | ぴゃ |
| | | | | | | | | | | | | | | | | | | | | | | | | | | | | | | | | | | | | | | | | | | | | | | | | | | | | | 0 | .08 | ぴゅ |
| | | | | | | | | | | | | | | | | | | | | | | | | | | | | | | | | | | | | | | | | | | | | | | | | | | | | | | 0 | ぴょ |